\documentclass[letterpaper, 10 pt, conference]{ieeeconf}  

\IEEEoverridecommandlockouts                              

\PassOptionsToPackage{table}{xcolor}

\usepackage{cite}
\usepackage{amsmath}
\usepackage{amssymb}
\usepackage{amsfonts}
\usepackage{graphicx}
\usepackage{textcomp}
\usepackage{xcolor}

\usepackage{url}
\usepackage{subcaption}
\usepackage{placeins} 
\usepackage{booktabs}
\usepackage{tabularx}
\usepackage{adjustbox}
\usepackage{xspace}
\usepackage{algorithm}
\usepackage[noend]{algpseudocode}
\usepackage{listings}

\let\labelindent\relax 
\usepackage{enumitem}

\usepackage{amsmath,amsfonts,bm}

\def\eqref#1{equation~\ref{#1}}

\def\1{\bm{1}}

\DeclareMathAlphabet{\mathsfit}{\encodingdefault}{\sfdefault}{m}{sl}
\SetMathAlphabet{\mathsfit}{bold}{\encodingdefault}{\sfdefault}{bx}{n}

\DeclareMathOperator*{\argmax}{arg\,max} 
\DeclareMathOperator*{\argmin}{arg\,min} 

\usepackage[bookmarks=true]{hyperref}
\usepackage{cleveref}
\hypersetup{pdfauthor={Arthur Bucker, Pramod Anantharam, Jonathan Francis, Jean Oh}, pdftitle={VG-Steer: Visual-Geometric Policy Steering from In-Context Aligned Abstractions}}

\newcommand{\ourmethod}{VG-Steer\xspace}

\newcommand{\cell}[5]{$#3$~{\scriptsize\textcolor{black!55}{$#1/#2$}}~{\scriptsize $[#4, #5]$}}
\newcommand{\cellb}[5]{$\mathbf{#3}$~{\scriptsize\textcolor{black!55}{$#1/#2$}}~{\scriptsize $[#4, #5]$}}

\definecolor{codebackground}{rgb}{1.0, 1.0, 1.0} 
\definecolor{codeblue}{rgb}{0.1, 0.5, 0.9}          
\definecolor{codegreen}{rgb}{0.1, 0.5, 0.1}         
\definecolor{codegray}{rgb}{0.5, 0.5, 0.5}          
\definecolor{codered}{rgb}{0.8, 0.1, 0.1}           
\definecolor{advisorgreen}{RGB}{136,189,117}
\definecolor{groundingyellow}{RGB}{255,224,133}
\definecolor{roboticblue}{RGB}{123,173,244}
\definecolor{monitorgray}{RGB}{183,183,183}

\lstdefinestyle{transcript}{
    basicstyle=\ttfamily\scriptsize,
    backgroundcolor=\color{codebackground},
    breaklines=true,
    breakatwhitespace=true,
    breakautoindent=true,
    columns=fullflexible,
    keepspaces=true,
    showstringspaces=false,
    frame=single,
    framesep=5pt,
    rulecolor=\color{codegray},
    xleftmargin=6pt,
    xrightmargin=2pt,
    aboveskip=4pt,
    belowskip=4pt,
    postbreak=,
    literate=%
        {–}{{-}}1 {—}{{---}}1 {…}{{\ldots}}1
        {°}{{\textdegree}}1 {≈}{{$\approx$}}1
        {‘}{{`}}1 {’}{{'}}1 {“}{{``}}1 {”}{{''}}1 {−}{{-}}1,
}

\newcommand{\xxnote}[3]{}
\ifx\hidenotes\undefined
\renewcommand{\xxnote}[3]{\color{#2}{#1: #3}}
\fi

\newcommand{\bbm}{\begin{bmatrix}}
\newcommand{\ebm}{\end{bmatrix}}

\title{\LARGE \bf
VG-Steer: Visual-Geometric Policy Steering\\from In-Context Aligned Abstractions
}

 \author{Arthur Bucker$^{1}$, Pramod Anantharam$^{2}$, Jonathan Francis$^{1,2}$ and Jean Oh$^{1,3}$
 \thanks{$^{1}$Arthur Bucker, Jonathan Francis and Jean Oh are with the Robotics
         Institute, Carnegie Mellon University, Pittsburgh, PA 15213, USA
         {\tt\small \{abucker,jmf1,jeanoh\}@cs.cmu.edu}}%
 \thanks{$^{2}$Pramod Anantharam and Jonathan Francis are with the Robot Learning Lab, Bosch Center for Artificial Intelligence, Pittsburgh, PA 15222, USA}%
 \thanks{$^{3}$Jean Oh is also with Lavoro AI, Inc., Pittsburgh, PA 15232, USA}%
 }

\begin{document}

\maketitle
\thispagestyle{empty}
\pagestyle{empty}

\begin{abstract}
Adapting robot policies to out-of-distribution (OOD) settings with limited real-world data remains a key bottleneck to generalist robots. The multimodality and scarcity of robotics data lead end-to-end approaches to overfit and break under small distribution shifts. Test-time adaptation can alleviate this by correcting frozen policies at inference, but most methods ground their correction in a signal learned implicitly from robotic datasets, making it susceptible to the same shifts as the policy it corrects. Recent methods that instead exploit explicit spatial information are typically limited to coarse corrections, missing the nuanced, task-relevant motion most manipulation tasks require. We propose \ourmethod, which corrects existing policies by transferring the nuanced motion of successful in-distribution rollouts, captured along meaningful geometric axes, known here as object-relational distance pairs.
\ourmethod leverages the semantic understanding of Large Vision Models (LVMs) and Vision-Language Models (VLMs) reasoning, which combines the task-relevant entity pairs and their object-relational distance projections into a single tailored visual-geometric guidance signal.
In a simulation benchmark, it efficiently corrects flow-matching policies under pose and task-goal shifts; on a real robot experiment, it widens the region of the workspace the frozen policy covers, more than doubling its success rate on cable placements one grid step outside the training positions, and steers a single-task policy toward objects it was never trained on while nearly eliminating object confusion. \textbf{Project website: \href{https://arthurfenderbucker.github.io/VG-Steer/}{https://arthurfenderbucker.github.io/VG-Steer/}}


\end{abstract}

\section{Introduction}
\label{sec:introduction}

Robotic manipulation policies trained at scale achieve strong in-distribution performance \cite{pi0, pi05, Gr00t, openvla, octo}, yet degrade under modest shifts in object appearance, geometry, layout, language instruction, or embodiment \cite{guo2025robustness, liberoplus, liberopro}. Retraining for every new setting is costly and data-hungry: collecting hundreds of demonstrations per task in each new environment is impractical.

Test-time adaptation offers an alternative: keep the policy fixed and use external correction signals to rerank its actions \cite{vgps, grappa, robomonkey, wu2025foresight, seal} or steer their generation \cite{zhang2025align, dynaguide, itps}. However, many steering approaches rely on learned value, reward, or latent signals that themselves fail under distribution shift, as reward models are prone to overoptimization on out-of-distribution inputs \cite{rewardopt}. Recent work introduces explicit spatial cues \cite{grappa, vls, omniguide}, but is typically limited to coarse, high-level corrections and does not reliably shape the fine-grained, task-relevant motion many manipulation tasks require.

\begin{figure*}[t]
    \centering
    \includegraphics[width=\linewidth]{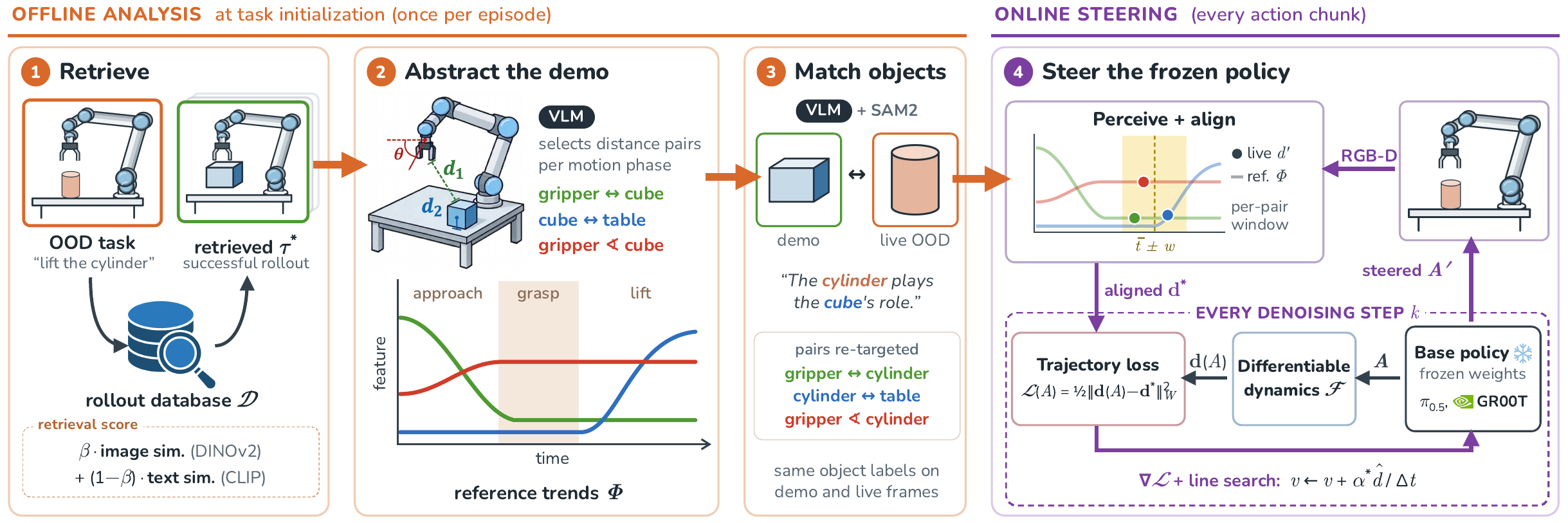} 
    \caption{\textbf{Overview of our visual-geometric policy steering framework.} \textit{Offline analysis (at task initialization, once per episode):} (1) we retrieve the most relevant successful in-distribution rollout $\tau^*$ by weighted image (DINOv2) and text (CLIP) similarity; (2) a VLM selects the task-relevant object-relational distance pairs per motion phase (e.g.,\ gripper--cube distance $d_1$, cube--table distance $d_2$, gripper orientation $\theta$), evaluated per timestep into the reference trends $\Phi$; (3) a VLM with SAM2 matches the demonstration objects to the live OOD scene (e.g.,\ the cylinder takes the cube's role), re-targeting each pair. \textit{Online steering (every action chunk):} (4) perception projects the live scene into the same distance space and aligns each pair to $\Phi$ within a window, giving the reference $\mathbf{d}^*$. At every denoising step, a differentiable dynamics model $\mathcal{F}$ forecasts the features $\mathbf{d}(A)$ of a candidate chunk $A$, and a line-searched gradient step on the trajectory loss corrects the frozen policy's velocity field, yielding the steered chunk $A'$.}
    \label{fig:overview}
    \vspace{-7mm}
\end{figure*}

This paper addresses that gap with \ourmethod, a visual-geometric steering framework that transfers behavior from retrieved in-distribution demonstrations to OOD scenes without updating policy weights. Our key idea is a single guidance signal built from object-relational distance pairs, evaluated per timestep along VLM-selected axes, so that it captures the nuanced motion of the retrieved rollout rather than only coarse endpoints. We (i) retrieve relevant demonstrations by visual and task similarity, (ii) extract object-relational distance-pair trajectories along automatically selected axes, and (iii) inject a differentiable guidance signal spanning position, orientation, and grasp directly into the policy's action-chunk generation (Figure~\ref{fig:overview}). The guidance is modular, test-time only, and compatible with pretrained flow-matching or diffusion action heads.

We evaluate on challenging OOD settings in LIBERO-Pro \cite{liberopro}. With $\pi_{0.5}$ \cite{pi05}, \ourmethod improves success under pose and task-goal shifts, outperforming VLS, our main training-free gradient-based steering baseline, and it transfers to GR00T N1.7 \cite{Gr00t} without retuning. We further deploy the complete pipeline, including open-vocabulary perception, on a physical UR5 under positional and task shift, where it widens the region of the workspace the frozen policy covers, more than doubling its success rate one grid step outside the training positions, and redirects a single-task policy toward objects it was never trained on, reducing object confusion by an order of magnitude. Our contributions are:
\begin{enumerate}
	\item A retrieval-to-steering pipeline that transfers behavior to OOD environments through explicit visual-geometric abstractions
	\item A distance-pair representation that captures the nuanced motion of a demonstration by tracking VLM-selected object relations per timestep along automatically chosen axes
	\item An empirical study on LIBERO-Pro and a UR5 deployment showing gains over the unsteered base and a state-of-the-art steering baseline without any weight update.
\end{enumerate}
\section{Related Works}
\label{sec:related_works}


\subsection{Policy Steering: Flow-Matching and Diffusion}

Policy steering adapts frozen generative policies at inference time through external guidance signals, avoiding costly retraining \cite{itps, dynaguide, vls}. Methods differ primarily in how they construct those signals.

\textbf{Implicit and gradient-free approaches.} Several works learn value or reward functions and use them to bias action sampling. VGPS \cite{vgps} computes a value-weighted score to rerank diffusion samples, and DynaFlow \cite{dynaflow} embeds physical dynamics into the flow-matching vector field. Effective in distribution, these methods struggle under OOD shifts where the learned models themselves stop providing reliable guidance, such as fine-grained or novel-object interactions absent from currently limited robotics datasets.

\textbf{Explicit spatial approaches.} A complementary line grounds guidance in semantically aligned geometric structures. OmniGuide \cite{omniguide} constructs guidance fields from structured scene representations, and GRAPPA \cite{grappa} and VLS \cite{vls} use explicit guidance functions over 3D geometric features. Despite their explicit spatial cues aligned with LVM semantic features, these methods target coarse, high-level corrections and lack mechanisms to align fine-grained contact and motion detail in OOD settings.

Our work bridges this gap with a single object-relational visual-geometric signal \cite{rekep} that, by tracking VLM-selected distance pairs per timestep along meaningful axes, aligns the nuanced motion of a demonstration rather than only its coarse endpoints. Steering the denoising process with this differentiable signal, like \cite{itps}, improves OOD generalization without modifying base policy weights.

\subsection{Visual-geometric representations}
Structured geometric representations have demonstrated the ability to enhance policy generalization. Wang et al. \cite{wang2022end} encode 3D keypoints for tool-object relations, and \cite{zhang2023universal} fuses spatial and semantic features for strong manipulation performance. HEPi \cite{hoang2025geometry} organizes robots and objects as graphs, but these methods typically rely on latent, non-transparent spaces.


\section{Method}
\label{sec:method}

Our goal is to steer a pretrained policy $\pi(A_t | o_t)$ to succeed in an out-of-distribution (OOD) target environment by leveraging relevant in-distribution rollouts $\mathcal{D} = \{\tau_i\}$. We propose \ourmethod, a modular, test-time framework that generates a single visual-geometric guidance signal, which transfers the nuanced motion of a retrieved successful rollout by tracking it along VLM-selected object-relational distance axes, and uses this signal to correct the policy's action generation process. An overview of \ourmethod is shown in Figure~\ref{fig:overview}.

\subsection{Problem Formulation}
\label{sec:problem_formulation}

We consider a pretrained generative policy $\pi(A_t \mid o_t)$ mapping an observation $o_t \in \mathbb{O}$ to an action chunk $A_t \in \mathbb{A}$, trained on an in-distribution environment and held \emph{fixed} throughout, and a database of successful in-distribution rollouts $\mathcal{D} = \{\tau_i\}_{i=1}^{|\mathcal{D}|}$ with $\tau_i = \{(o_0, A_0), \ldots, (o_H, A_H)\}$. At test time the robot operates in an OOD environment whose observations $o'_t$ exhibit shifts in appearance, object geometry, layout, language instruction, or embodiment.

\textbf{Goal.} Without updating $\pi$, generate corrected action chunks $A'_0, \ldots, A'_T$ that transfer the task-relevant behavior from $\mathcal{D}$ to the OOD scene. We distill the retrieved demonstration $\tau^* \in \mathcal{D}$ into a compact, task-tailored guidance signal: a set of object-relational distance pairs that evolve over time and encode an efficient heuristic for completing the task. This time-varying signal then steers the policy's action generation at inference.

\subsection{Background: Steering the Denoising Process}
\label{sec:background}

The policies we target synthesize an action chunk by iteratively denoising a noisy sample, either by integrating a learned velocity field (flow matching) or by following a learned score (diffusion). Such a frozen policy can be \emph{steered} at inference, without any weight update, by perturbing each denoising step with the gradient of an auxiliary guidance objective: an idea originating in classifier guidance \cite{classifierguidance}, generalized to arbitrary differentiable objectives by training-free universal guidance \cite{universalguidance}, and used by diffusion planners to bias sampling toward high-return plans \cite{diffuser}. For flow matching, the correction is applied to the velocity field rather than the score.

Recent robot-policy work adopts this mechanism, differing mainly in \emph{what} guidance signal is constructed: human interactions \cite{itps}, learned dynamics or value models \cite{dynaguide, vgps}, or explicit spatial energies \cite{omniguide, vls, grappa}. Our contribution is thus not the steering machinery, which we inherit, but the guidance \emph{signal}: a visual-geometric, object-relational distance-pair objective distilled from a temporally aligned retrieved demonstration (Sections~\ref{sec:vg_guidance} and \ref{sec:guidance_injection}).

\subsection{Demonstration Retrieval and Abstraction}
\label{sec:retrieval}

\textbf{Retrieval.} Given the current OOD task defined by initial observation $o'_0$ and language instruction $\ell$, we retrieve the most relevant demonstration by combining a visual and a language similarity as a weighted average:
\begin{equation}
\begin{split}
    \tau^* = \argmax_{\tau_i \in \mathcal{D}} \; &\beta \, \mathrm{sim}\!\left(\psi_v(o'_0),\, \psi_v(\tau_i[0])\right) \\
    &+ (1-\beta) \, \mathrm{sim}\!\left(\psi_\ell(\ell),\, \psi_\ell(\ell_i)\right),
\end{split}
\end{equation}
where $\mathrm{sim}(a, b) = \cos(a, b)$ is the cosine similarity between $\ell_2$-normalized embeddings, $\psi_v$ embeds the initial RGB observation with a DINOv2 image encoder \cite{dinov2} (averaged over frames), and $\psi_\ell$ embeds the language instruction with a CLIP text encoder \cite{clip}. The weight $\beta$ trades off visual against language similarity and is held fixed across experiments.

\textbf{Abstraction.} The retrieved demonstration $\tau^*$ is processed into an \emph{object-relational distance-pair trajectory} $\Phi$ encoding the geometric relations between the end-effector (EEF) and task-relevant objects (Section~\ref{sec:vg_guidance}). Because these relations are evaluated per timestep along VLM-selected axes, $\Phi$ captures how the demonstrated motion unfolds along the task-relevant axes rather than only its coarse endpoints.



\subsection{Visual-Geometric Guidance via Object-Relational Distance Pairs}
\label{sec:vg_guidance}

We represent task structure as a set of pairwise geometric relations between task-relevant entities, where both the relevant entities \emph{and} the appropriate distance projection are determined automatically per sub-trajectory phase. Rather than a single coarse constraint, evaluating these relations at every timestep yields a time-resolved profile that captures the nuanced motion of the reference rollout along the meaningful axes the VLM selects (Appendix~\ref{sec:appendix-demo-analysis}).

\textbf{Sub-trajectory decomposition.} We segment $\tau^*$ into $M$ semantically coherent phases $\{ST_m\}_{m=0}^{M-1}$ following the velocity-based heuristic of STRAP \cite{strap}, splitting at low end-effector velocity points. Each phase's end-effector path is rendered as a distinct colored overlay on the demo frame, giving the VLM a visual summary of how the motion unfolds.

\textbf{Scene segmentation.} We obtain object instance masks $\{(s_j, \text{name}_j)\}$ with an open-vocabulary pipeline: a VLM (Gemini~3~Flash \cite{gemini}) proposes a grounded bounding box per task-relevant object and SAM2 \cite{sam2} converts each box into a precise mask, with $s_j$ the instance ID and $\text{name}_j$ its label. Running the same pipeline on the demonstration frame and the live OOD frame gives the two scenes a consistent language-grounded labeling, and yields the segmentation- and sub-trajectory-overlay images fed to the VLM.

\textbf{VLM-guided distance pair selection.} A VLM is queried with the task description, the original RGB frame, the segmentation-overlay image, and the sub-trajectory-overlay image. For each sub-trajectory phase $ST_m$ the VLM outputs a set of entity pairs $\mathcal{P}_m$, each annotated with a \textit{relation type} and an optional projection:
\begin{itemize}[nosep]
    \item \textbf{positional}: a \textbf{centroid} (Euclidean distance between 3D centroids) or \textbf{surface} (shortest source-to-surface distance, used when contact or proximity is critical) distance, with an optional \textit{projection} that keeps only the signed component along an \textbf{axis} $\hat{u}\in\mathbb{R}^3$ or the in-plane component w.r.t.\ a \textbf{plane} normal $\hat{n}\in\mathbb{R}^3$ (null retains the full 3D distance);
    \item \textbf{orientation}: the alignment of an end-effector body-frame vector $R\,\hat{e}$ with a selected axis (a signed cosine) or plane (the residual in-plane magnitude), constraining pose rather than position;
    \item \textbf{gripper}: the gripper aperture, capturing grasp timing.
\end{itemize}
The source of a positional/orientation relation is the end-effector, a fingertip, or a grasped object, and the paired entity is a segmented object identified by its instance ID. Each phase thus receives its own relation set, grounded in the visual content of the demonstration, with position, orientation, and grasp in a single common form.

\textbf{Geometric feature trajectory.} For phase $m$ and timestep $t \in ST_m$, each relation $n \in \mathcal{P}_m$ is reduced to a scalar feature:
\begin{equation}
    d^{n}_t = \Pi_n\!\left( \delta^{n}_t \right), \quad
    \phi^t = \left[ d^{n}_t \right]_{n\in\mathcal{P}_{m(t)}} \in \mathbb{R}^{|\mathcal{P}_{m(t)}|},
\end{equation}
where $\delta^{n}_t$ is the raw geometric quantity of relation $n$ (a centroid/surface distance vector, a body-frame orientation vector, or the gripper aperture) and $\Pi_n(\cdot)$ is its read-out: the configured axis/plane projection or identity for positional relations, the signed-cosine/in-plane projection for orientation relations, and the identity for gripper relations. Object centroids and surface points are obtained by unprojecting the segmentation mask with the depth image. This defines the visual-geometric trajectory $\Phi = \{\phi^t\}_{t=0}^{H}$.

\textbf{Decoding an action chunk to geometric features.} In simulation, the base policies we steer emit operational-space-control (OSC) action chunks: per-step end-effector linear and angular velocities plus a gripper command. To evaluate the relations on a candidate chunk $A_t$, a differentiable dynamics model $\mathcal{F}$ rolls it out from the current state $s_0 = (p_0, R_0, g_0)$,
\begin{equation}
    \{(p_{t'}, R_{t'}, g_{t'})\}_{t'=1}^{T} = \mathcal{F}(A_t,\, s_0),
\end{equation}
predicting end-effector position, orientation, and gripper aperture at each chunk step $t'$; fingertip and held-object source points follow by forward kinematics. Combined with the object centroids and surface points read from the current observation, this turns a candidate chunk into a \emph{forecast} of the distance pairs it would induce: a predicted feature $d^{n}_{t'}(A_t)$ for every relation $n$ and step $t'$, under the same pairs $\mathcal{P}_{m^*}$ and projections selected on the demonstration. The whole read-out is differentiable and JIT-compilable, adding only a lightweight forward pass, so it can be evaluated at every denoising step. We implement $\mathcal{F}$ as a kinematic integrator of the OSC velocities for the simulated experiments (Appendix~\ref{sec:appendix-rollout}). In the real world, the policy acts in joint positions, so $\mathcal{F}$ is simply the UR5 forward kinematics.

\textbf{Visual-geometric guidance loss.} Let $\{d^{*n}_{t'}\}$ be the reference distance trajectory of the retrieved demonstration, aligned to the robot's current progress (Section~\ref{sec:guidance_injection}). The guidance objective penalizes the per-relation, per-step deviation from this reference across the entire chunk:
\begin{equation}
    \mathcal{L}_{guide}(A_t) = \tfrac{1}{2}\sum_{n=1}^{N} \sum_{t'=1}^{T} w_n \left( d^{n}_{t'}(A_t) - d^{*n}_{t'} \right)^2,
    \label{eq:guidance_loss}
\end{equation}
where $N$ is the number of active relations, $T$ the chunk horizon, and $w_n \ge 0$ a per-relation weight combining the VLM-assigned importance with a relevance gate that activates each relation only as the end-effector approaches the relevant object (Section~\ref{sec:guidance_injection}). Since orientation and gripper relations enter the same sum, a single objective jointly captures position, orientation, and grasp timing along the whole predicted chunk rather than at a single waypoint.

\subsection{Differentiable Guidance Injection}
\label{sec:guidance_injection}

\textbf{Trajectory alignment.} The reference targets $d^{*n}_{t'}$ in Eq.~\ref{eq:guidance_loss} are obtained by aligning the live execution to the demonstration in the feature space. Instead of matching all relations to one shared frame, we align \emph{each relation independently} within a bounded window around a shared cursor $\bar{t}$ (the MIXED scheme in our implementation): for relation $n$ with current live value ${d'}^{n}$,
\begin{equation}
    t^*_n = \argmin_{|t - \bar{t}| \le w} \bigl( {d'}^{n} - d^{n}_t \bigr)^2,
\end{equation}
and we read $d^{*n}_{t'} = d^{n}_{t^*_n + t'}$ as that relation's per-step references. The cursor is \emph{not} constrained to advance monotonically: within the active sub-trajectory it may move forward or backward by at most a fixed number of steps per update, so the guidance can follow corrective, retry behavior (e.g.\ re-approaching an object after a missed grasp) rather than being dragged through stages the robot has not completed. Sub-trajectory switching is likewise bidirectional and gated on sub-goal satisfaction, analogous to the stage-wise relational keypoint constraints of ReKep \cite{rekep}: the active relation set $\mathcal{P}_{m}$ advances when a phase's sub-goal is met and falls back if it no longer holds. Window sizes and switching hysteresis are in Appendix~\ref{sec:appendix-alignment}.

\textbf{Orientation without rotation-representation artifacts.} Euclidean distances over rotation parameters cause singularities and trajectory artifacts. Rather than add a separate rotation loss, we constrain orientation \emph{inside} $\mathcal{L}_{guide}$ through the orientation relations of Section~\ref{sec:vg_guidance}, each projecting a rotated end-effector body-frame vector $R_{t'}\hat{e}$ onto a VLM-selected axis or plane. Because orientation is propagated as a rotation and only its projection enters the loss, the objective stays well-behaved on $SO(3)$ while remaining a plain distance-pair term.

\textbf{Line-searched guidance correction.} We inject guidance at every denoising step $k$, without modifying policy weights. Let $A^{(k)}_t$ be the current noisy chunk and $v$ the policy's predicted velocity field (for diffusion, the score-derived update). We form a unit descent direction from the guidance gradient at the policy's tentative next iterate,
\begin{equation}
    g = \nabla_{A}\,\mathcal{L}_{guide}\!\left(A^{(k)}_t + \Delta t\, v\right), \qquad
    \hat{d} = -\,g/\lVert g\rVert,
\end{equation}
so the correction magnitude is set by a step size rather than the raw gradient scale. A single global step is brittle, as early high-noise steps tolerate only small corrections while later ones benefit from larger ones, so we select the step \emph{per denoising step} with a one-dimensional line search,
\begin{equation}
    \alpha^\star = \argmin_{\alpha \in [0,\,\alpha_{\max}]} \mathcal{L}_{guide}\!\left(A^{(k)}_t + \alpha\,\hat{d}\right),
\end{equation}
solved by a golden-section search \cite{goldensection} with a \emph{do-no-harm} guard: if no positive step reduces the loss, $\alpha^\star = 0$ and the unsteered base step is kept (Algorithm~\ref{alg:line_search}). The step is injected into the velocity field,
\begin{equation}
    v \leftarrow v + \frac{\alpha^\star}{\Delta t}\,\hat{d}\,c,
    \label{eq:velocity_correction}
\end{equation}
so the sampler update becomes $A^{(k+1)}_t = A^{(k)}_t + \Delta t\,v + c\,\alpha^\star\,\hat{d}$: the base step plus a descent step of length $c\,\alpha^\star$ along $\hat{d}$, where $c$ is a global guidance scale. This reshapes the velocity field of a flow-matching policy, or the score estimate of a diffusion one, analogous to classifier guidance \cite{classifierguidance}, while the policy weights stay frozen.

\textbf{Action-space mapping.} In simulation, the policies we evaluate emit OSC delta actions in the end-effector Cartesian frame, so $\nabla_{A_t} \mathcal{L}_{guide}$ is already expressed in the policy's action space and the correction is applied without kinematic remapping. For the real-world experiment, the policies operate in joint-space, and the same Cartesian gradient maps into the action space through the pseudo-inverse Jacobian, $\delta A_t = J(q_t)^{\dagger}\, \nabla_{e_t} \mathcal{L}_{guide}$, analogous to how dynamics-aware flow-matching methods enforce embodiment-consistent corrections \cite{dynaflow}.

\begin{figure}[h]
    \centering
    \vspace{-3mm}
    \includegraphics[width=\linewidth]{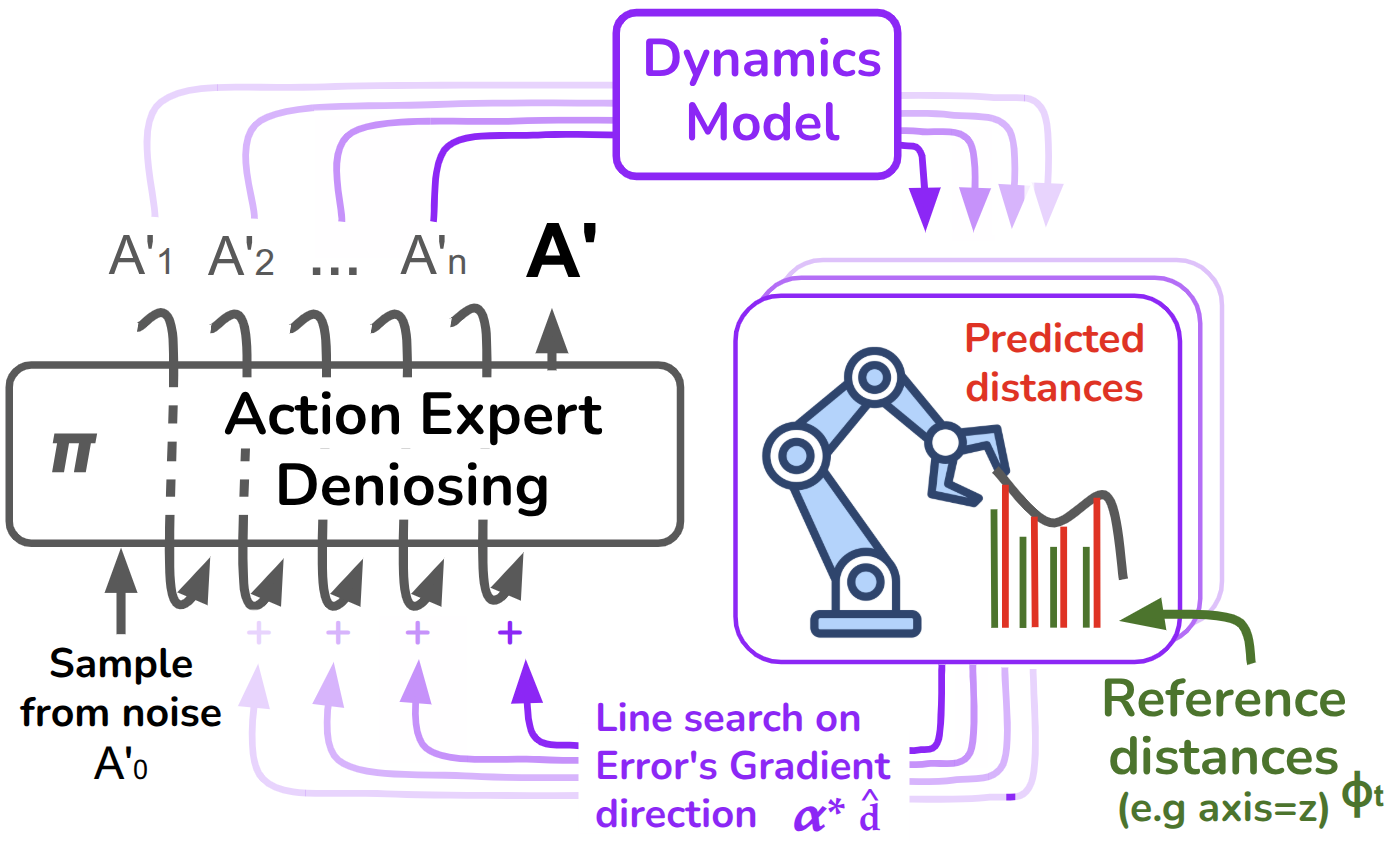}
    \caption{Differentiable guidance injection. At each denoising step, the candidate chunk is rolled out by forward kinematics, and the guidance loss (position, orientation, gripper) is differentiated; a golden-section line search sets the step size, and the correction is applied to the policy's velocity field, steering the trajectory without modifying weights.}
    \label{fig:steering_loop}
    \vspace{-5mm}
\end{figure}


\section{Implementation Details}
\label{sec:implementation-details}

\textbf{Base policies.} We evaluate on two pretrained flow-matching manipulation policies: $\pi_{0.5}$ \cite{pi05}, a vision-language-action policy pretrained on a large multi-embodiment dataset, and GR00T~N1.7 \cite{Gr00t}, a cross-embodiment foundation policy (its LIBERO checkpoint). Both are kept frozen and only the test-time guidance signal is applied during denoising.

\textbf{Simulation benchmark.} We report simulation results on LIBERO\_PRO \cite{liberopro}, an OOD extension of LIBERO introducing pose, object, instruction, and task-goal shifts.

\textbf{Guidance hyperparameters.} At each denoising step the guidance scale $\alpha^\star$ is chosen by the golden-section line search (Algorithm~\ref{alg:line_search}), whose do-no-harm guard always admits the unsteered base step ($\alpha^\star = 0$). The global scale $c$ and the relevance-gating thresholds setting $w_n$ in Eq.~\ref{eq:guidance_loss} are constant across all benchmarks and policies; only the search bound $\alpha_{\max}$ is rescaled per policy for its action parameterization (velocity field versus score). The remaining hyperparameters (retrieval weight $\beta$, STRAP thresholds, VLM prompts) are shared across experiments. Full values and the perception, alignment, and weighting details are in Appendix~\ref{sec:appendix-hyperparams}.

\section{Simulation Experiments}
\label{sec:results}

We evaluate on LIBERO-Pro across two base policies, measuring task success rate under multiple types of distribution shift, against VLS \cite{vls}, our main training-free gradient-based steering baseline.

\textbf{Q1: Does visual-geometric steering improve OOD generalization over existing policy steering baselines?} We center the comparison on the \emph{task} and \emph{positional} perturbations, the regime where the base policies are weakest and the only one the VLS protocol evaluates. All methods in Table~\ref{tab:main_results}, including VLS, receive the same simulator-provided object segmentation and labels, as in the released VLS LIBERO evaluation code, so the comparison isolates the steering signal from perception (real open-vocabulary perception in Appendix~\ref{sec:appendix-realperception}).

\begin{table}[!ht]
    \centering
    \caption{Success rate (\%) on the task and positional perturbations of LIBERO-Pro ($\pi_{0.5}$), each averaged over the four suites, and their mean (\emph{Average}). All rows use simulator-provided object segmentation and labels. $^\ast$VLS reproduced in our setup using the released VLS codebase (\emph{grad-only} disables its population resampling); see Appendix~\ref{sec:appendix-vls}.}
    \label{tab:main_results}
    \begin{tabular}{lccc}
        \toprule
        Method ($\pi_{0.5}$) & Task & Pos & Average \\
        \midrule
        No steering                   & 26.3 & 26.3 & 26.3 \\
        VLS (reproduced, full)$^\ast$ & 32.0 & 27.6 & 29.8 \\
        VLS (grad-only)$^\ast$        & 17.3 & 21.5 & 19.4 \\
        \ourmethod                     & \textbf{40.8} & \textbf{34.3} & \textbf{37.5} \\
        \bottomrule
    \end{tabular}
\end{table}

\ourmethod raises the task success rate from $26.3$ to $40.8$ and the positional rate from $26.3$ to $34.3$ ($37.5$ average against $26.3$), surpassing full VLS ($29.8$). The gap is starkest against the compute-matched grad-only VLS: without population resampling it falls to $19.4$, \emph{below} the unsteered base, whereas the single-chain gradient guidance of \ourmethod still improves by a wide margin. As both are gradient-only, the difference is attributable to the guidance \emph{signal} rather than to VLS's sampling machinery. Conversely, where the base policy is already near-saturated (object and language, $75$--$95\%$) there is little gain, and the guidance can even be \emph{detrimental}, perturbing an already-correct trajectory; the relevance gate (Section~\ref{sec:guidance_injection}) mitigates but does not eliminate this. The per-type breakdown is in Appendix~\ref{sec:appendix-breakdown}.


\begin{table}[!t]
    \centering
    \caption{Transfer across base policies (LIBERO-Pro; task and positional success rate \%, averaged over suites, and their mean).}
    \label{tab:cross_policy}
    \begin{tabular}{llccc}
        \toprule
        Base policy & Steering & Task & Pos & Average \\
        \midrule
        $\pi_{0.5}$ (flow)   & Base   & 26.3 & 26.3 & 26.3 \\
                             & + \ourmethod & \textbf{40.8} & \textbf{34.3} & \textbf{37.5} \\
        \addlinespace
        GR00T N1.7 (flow)    & Base   & 19.3 & 0.5  & 9.9 \\
                             & + \ourmethod & \textbf{20.0} & \textbf{3.5}  & \textbf{11.8} \\
        \bottomrule
    \end{tabular}
\end{table}

\noindent\textit{Transfer across base policies.} The same guidance plugs into a second flow-matching policy, GR00T~N1.7, without retuning (only $\alpha_{\max}$ is rescaled). It improves the task-and-positional average of both policies (Table~\ref{tab:cross_policy}): $26.3\!\to\!37.5$ for $\pi_{0.5}$ and $9.9\!\to\!11.8$ for GR00T~N1.7. The smaller GR00T gain reflects its base: at $0.5\%$ on positional shifts, there is little competent behavior to steer. Steering reshapes what the policy samples rather than supplying a missing skill, so gains scale with base-policy competence, consistent with the far-OOD dead band of the real-world sweep (Section~\ref{sec:real_world}).


\begin{figure}[!b]
    \centering
    \includegraphics[width=0.66\linewidth]{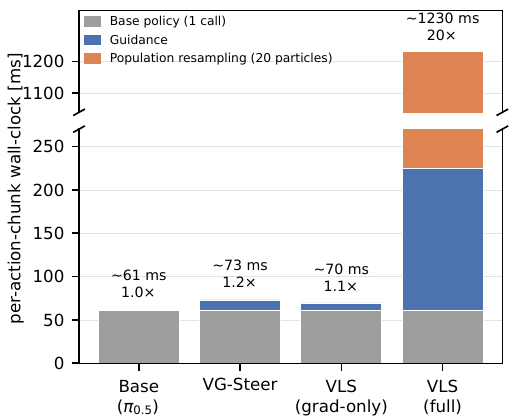}
    \caption{Per-action-chunk inference overhead on $\pi_{0.5}$ (mean over the four suites, CUDA-synchronized). With \ourmethod a chunk takes $\sim$73 ms in total ($1.2\times$ the base's $\sim$61 ms), comparable to gradient-only VLS ($\sim$70 ms total); full VLS costs $\sim$20$\times$ more, scaling with its particle count ($20$ here).}
    \label{fig:profiling}
    \vspace{-5mm}
\end{figure}

\textbf{Q2: What is the inference-time overhead of guidance?} Unlike VLS, which resamples a population of particles at every denoising step, \ourmethod runs a single denoising chain and adds only a lightweight, JIT-compiled distance-pair loss during the per-step line search (no extra policy forward pass). On $\pi_{0.5}$ this is $\sim$$12$\,ms over a single base-policy call ($\sim$$61$\,ms), a $1.2\times$ overhead on par with gradient-only VLS ($1.1\times$), whereas full VLS is $\sim$$20\times$ slower because its resampling re-runs the policy across particles (Figure~\ref{fig:profiling}). On hardware, perception adds $113\pm2$\,ms per inference plus a one-time first-frame setup of $29.5\pm3.1$\,s (Appendix~\ref{sec:appendix-realtiming}).

\textbf{Q3: Which components of the guidance mechanism drive the gains?} Turning the raw gradient of distance-pair errors into a usable correction relies on two mechanisms: the per-step line search that sets the step size (and rejects steps that increase the loss), and the normalization of the guidance gradient to a unit direction, which decouples the step from the raw gradient scale (Appendix~\ref{sec:appendix-injection}). We enable each cumulatively on top of the unsteered $\pi_{0.5}$ base (Table~\ref{tab:ablation}).

Raw fixed-step gradient guidance is slightly worse than the base ($23.8$ vs.\ $26.3$ average): unnormalized distance gradients are poorly scaled, so a fixed step over- or under-shoots. The per-step line search recovers a large gain ($23.8\!\to\!35.0$), and gradient normalization yields the full method and the best result ($35.0\!\to\!37.5$). The geometric signal alone is thus necessary but not sufficient: the line search and normalization are what make it a net-positive correction.


\begin{table}[h]
    \centering
    \caption{Guidance-mechanism ablation ($\pi_{0.5}$, LIBERO-Pro; success rate \%, averaged over the four suites). Components are enabled cumulatively on top of the unsteered base. The two intermediate variants use a $3$-trial sweep ($30$ episodes/cell); the base and full method use the $10$-trial main runs.}
    \label{tab:ablation}
    \begin{tabular}{lccc}
        \toprule
        Variant ($\pi_{0.5}$) & Task & Pos & Average \\
        \midrule
        No guidance             & 26.3 & 26.3 & 26.3 \\
        + gradient guidance (fixed step) & 24.2 & 23.3 & 23.8 \\
        + line search                   & 34.2 & \textbf{35.8} & 35.0 \\
        + gradient normalization (\ourmethod) & \textbf{40.8} & 34.3 & \textbf{37.5} \\
        \bottomrule
    \end{tabular}
\end{table}

\section{Real-World Experiments}
\label{sec:real_world}

We deploy the \emph{complete} pipeline on hardware, where grounding, depth,
calibration and control are all subject to real noise, and ask whether the
guidance still recovers the base policy's OOD failures. All real-world runs use
the open-vocabulary perception stack of
Appendix~\ref{sec:appendix-perception} (no privileged labels).


\textbf{Hardware.} A UR5 arm with a Robotiq~2F-140 parallel gripper. Object
detection and 3D reconstruction are driven by a single fixed external RealSense
D435, calibrated markerlessly with Kalib \cite{kalib}. Policy inference, the VLM
calls and the guidance loop all run on one RTX~5090
(Appendix~\ref{sec:appendix-realworld}).

\begin{figure}[!h]
    \centering
    \begin{subfigure}[b]{0.49\columnwidth}
        \centering
        \adjustbox{trim={0.10\width} {0} {0.0} {0.28\height},clip,width=\linewidth}{\includegraphics{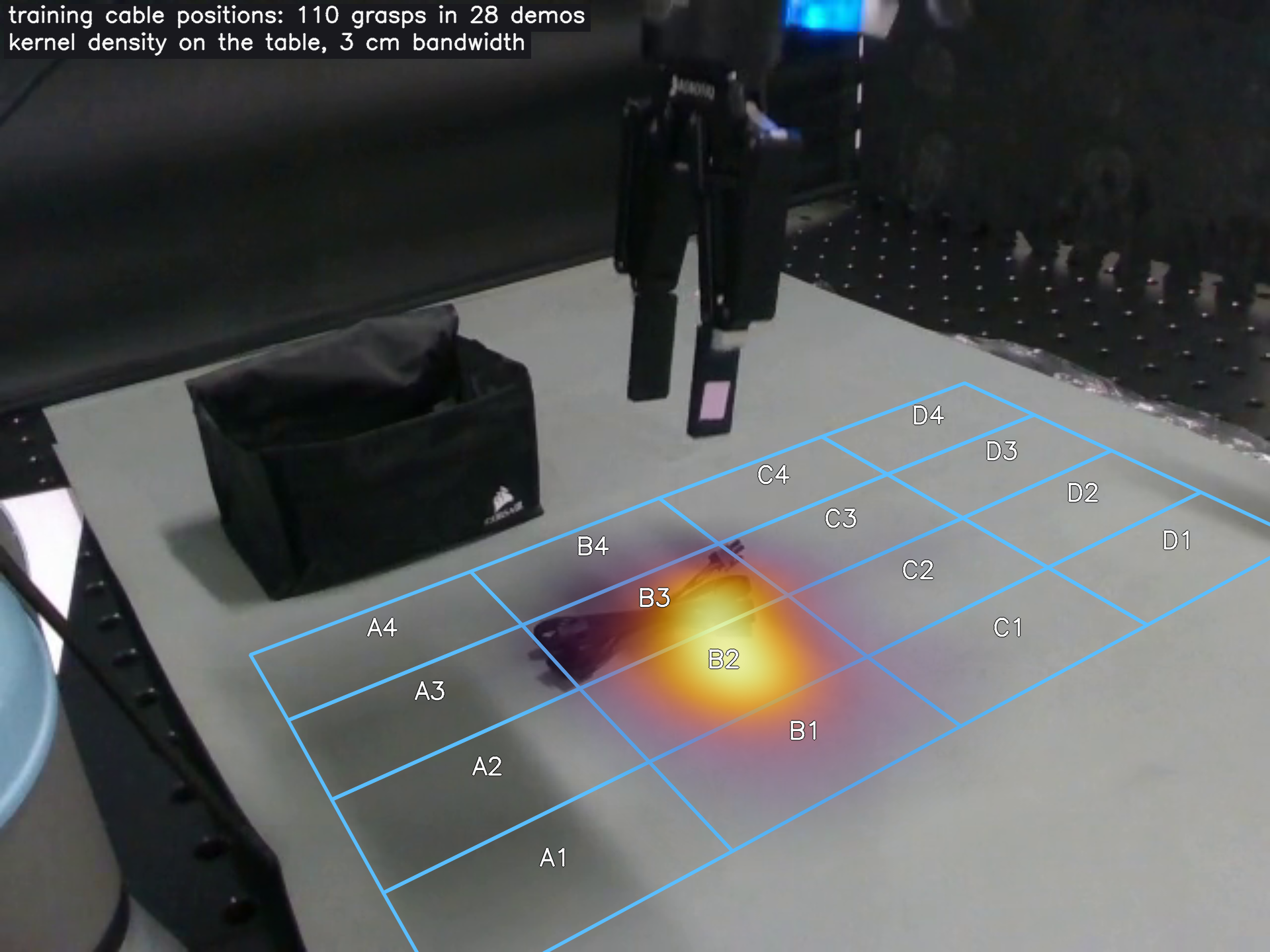}}
        \caption{Training Distribution.}
        \label{fig:real_positional_training}
    \end{subfigure}\hfill
    \begin{subfigure}[b]{0.49\columnwidth}
        \centering
        \adjustbox{trim={0.10\width} {0} {0} {0.28\height},clip,width=\linewidth}{\includegraphics{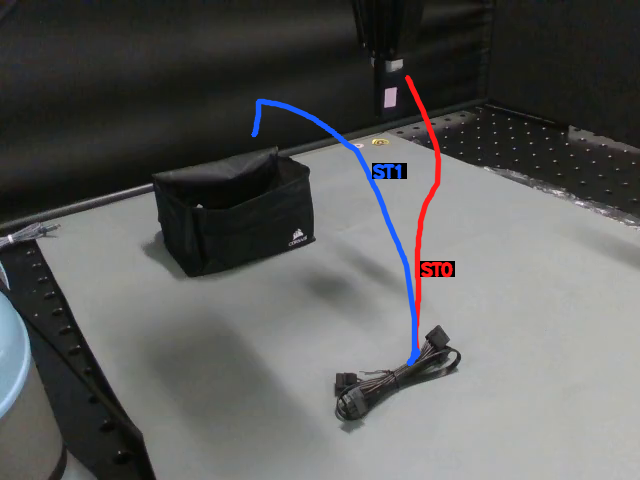}}
        \caption{Reference rollout.}
        \label{fig:real_positional_reference}
    \end{subfigure}

    \vspace{0.6em}
    \begin{subfigure}[b]{0.49\columnwidth}
        \centering
        \adjustbox{trim={0.10\width} {0} {0} {0.38\height},clip,width=\linewidth}{\includegraphics{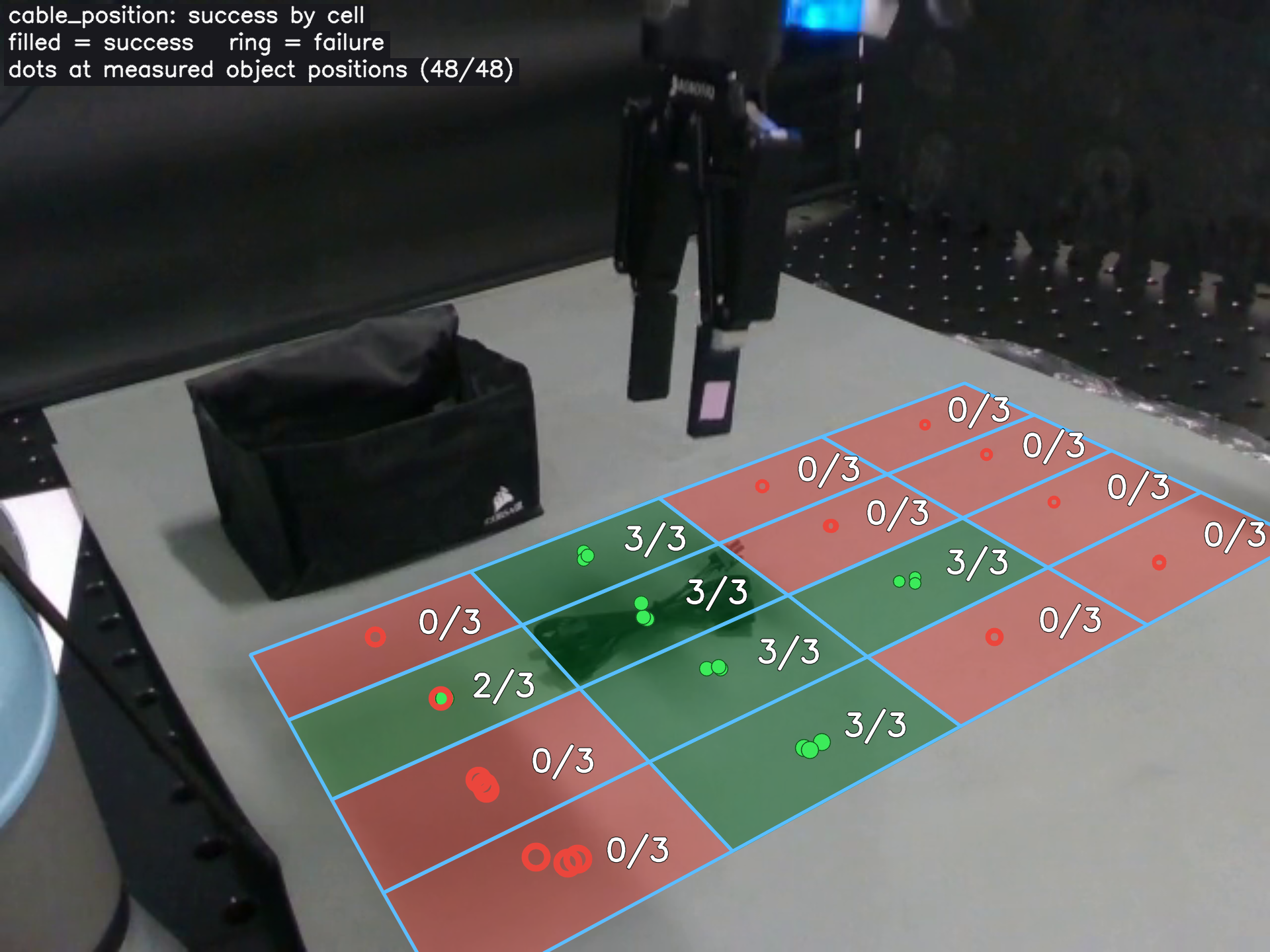}}
        \caption{Base (no steering).}
        \label{fig:real_positional_grid_base}
    \end{subfigure}\hfill
    \begin{subfigure}[b]{0.49\columnwidth}
        \centering
        \adjustbox{trim={0.10\width} {0} {0} {0.38\height},clip,width=\linewidth}{\includegraphics{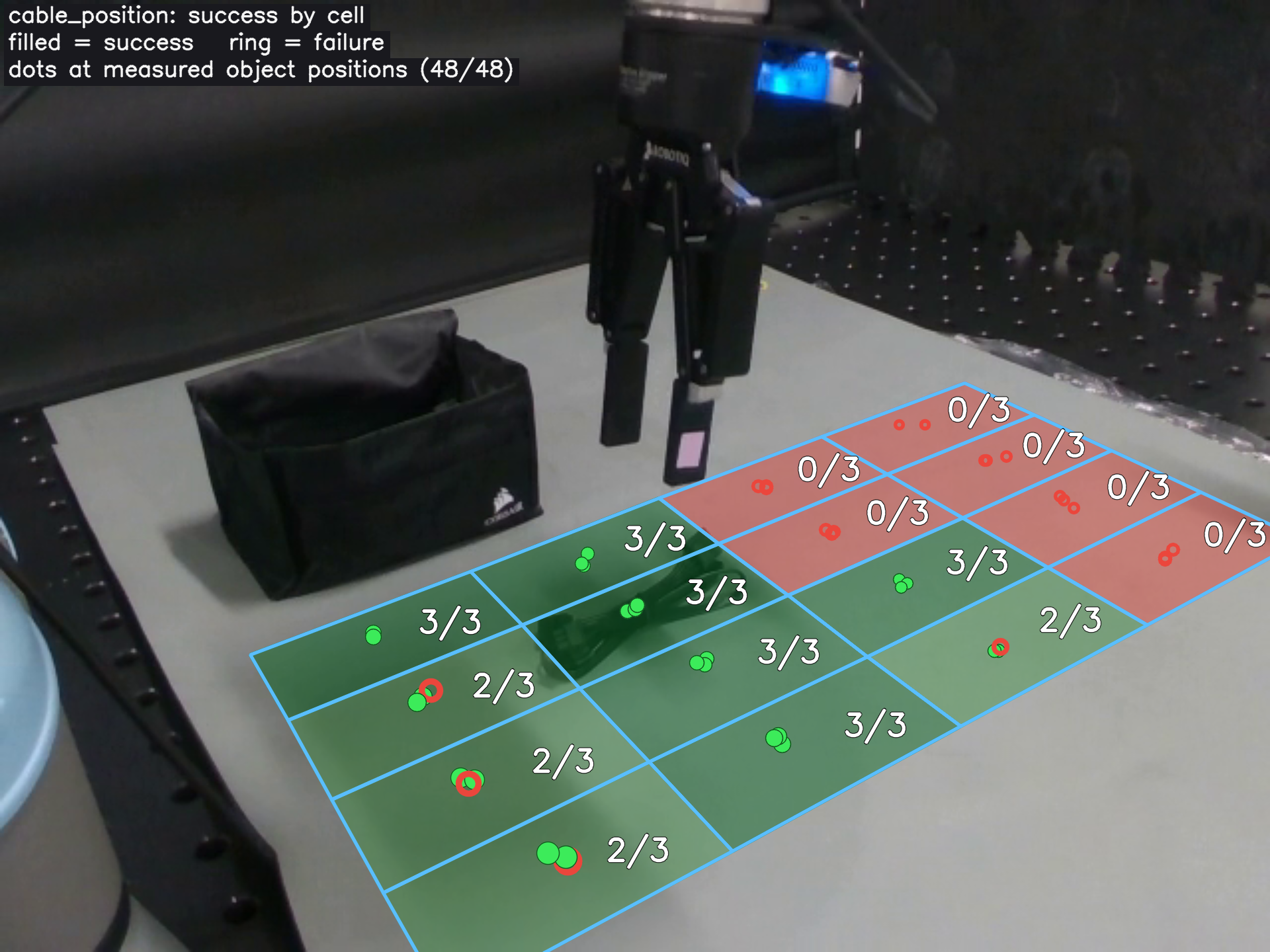}}
        \caption{\ourmethod.}
        \label{fig:real_positional_grid_ours}
    \end{subfigure}
    \caption{Positional sweep on cable bagging, in the camera view. (a) Training
    distribution of the $110$ cable grasps in the demonstrations. (b) The single reference rollout,
    with its end-effector path split into sub-trajectories. (c, d) Per-cell
    outcomes: dots mark the measured cable position of every trial, filled green
    for success, red rings for failure.}
    \label{fig:real_positional_grid}

\end{figure}

\textbf{Task and base policy.} We evaluate variations of a \emph{cable bagging}
task with a single $\pi_{0.5}$ \cite{pi05} checkpoint, finetuned on $28$
demonstrations of ``Put all cables in the bag'' and frozen throughout. It takes
one wrist and one fixed camera view and predicts joint positions
(Figure~\ref{fig:real_positional_grid}). The demonstration database
$\mathcal{D}$ holds a \emph{single successful rollout of that same policy} in
the nominal configuration: no human demonstrations, no extra training data and
no weight updates, so any OOD gain comes from how that one rollout is abstracted
and re-applied.

\begin{itemize}[nosep,leftmargin=*]
    \item \textbf{Positional variation.} Using the ``cable bagging'' policy, we
    vary the cable placement over a $4\times4$ grid, 3 trials per cell. Only
    three of the sixteen cells overlap the positions seen in training; the rest
    lies between two and six standard deviations away
    (Section~\ref{sec:real_positional}).
    \item \textbf{Task variation.} Using the ``cable bagging'' policy (trained only on cables), we vary the task by changing the name of the target object to be bagged: ``cable'', ``banana'', or ``spoon''. The three elongated objects are placed simultaneously inside an in-distribution region of the table
    (Section~\ref{sec:real_objects}).
\end{itemize}


\noindent The guidance improves both axes ($35.4\!\to\!54.2$ positional, $33.3\!\to\!62.5$
task), and the two failures it corrects differ in kind: under positional shift
the base policy knows \emph{what} to grasp and fails to reach it; under task
shift it reaches confidently, but for the \emph{wrong} object. 

\textbf{Q4: Does the guidance recover positional OOD failures?} \label{sec:real_positional}%
We run $48$ placements per method, three per cell of a $4\times4$ grid spaced
$15$ and $7.5$\,cm, re-placing the cable at the same nominal cell for both
methods (Figure~\ref{fig:real_positional_grid}). To put every placement on one
scale, we fit a Gaussian to the $110$ cable grasp positions in the $28$ training
demonstrations ($5.6$ and $6.2$\,cm along the two table axes) and report each
rollout's measured start position as a Mahalanobis distance in $\sigma$. We cut
three zones at $2$ and $4.5\sigma$; both cuts fall in empty gaps of the measured
distances (Appendix~\ref{sec:appendix-positional}).

\begin{figure}[!t]
    \centering
    \begingroup
    \captionsetup{type=table}
    \caption{Positional sweep on cable bagging, $48$ placements per method,
    every placement run by both. Zones are Mahalanobis bands cut at $2$ and
    $4.5\sigma$. Cells give the success rate (\%), successes/rollouts and the
    $95\%$ Wilson interval; $\Delta$ is the gain in points and $p$ a two-sided
    Fisher exact test.}
    \label{tab:real_positional}
    \adjustbox{max width=\columnwidth}{
    \begin{tabular}{@{}lcccc@{}}
        \toprule
        & \multicolumn{2}{c}{Success rate (\%)} & & \\
        \cmidrule(lr){2-3}
        Zone & Base & \ourmethod & $\Delta$ & $p$ \\
        \midrule
        In-distribution ($<2\sigma$)
            & \cell{9}{9}{100.0}{70}{100} & \cell{9}{9}{100.0}{70}{100}
            & $0.0$ & $1.00$ \\
        Near-OOD ($2$ to $4.5\sigma$)
            & \cell{8}{27}{29.6}{16}{48} & \cellb{17}{27}{63.0}{44}{79}
            & $\mathbf{+33.3}$ & $\mathbf{0.028}$ \\
        Far-OOD ($>4.5\sigma$)
            & \cell{0}{12}{0.0}{0}{24} & \cell{0}{12}{0.0}{0}{24}
            & $0.0$ & $1.00$ \\
        \midrule
        \textit{Pooled OOD} ($>2\sigma$)
            & \cell{8}{39}{20.5}{11}{36} & \cellb{17}{39}{43.6}{29}{59}
            & $+23.1$ & $0.051$ \\
        \textit{Pooled, full sweep}
            & \cell{17}{48}{35.4}{23}{50} & \cellb{26}{48}{54.2}{40}{67}
            & $+18.8$ & $0.10$ \\
        \bottomrule
    \end{tabular}
    }
    \endgroup

    \vspace{1.4em}
    \includegraphics[width=0.92\columnwidth]{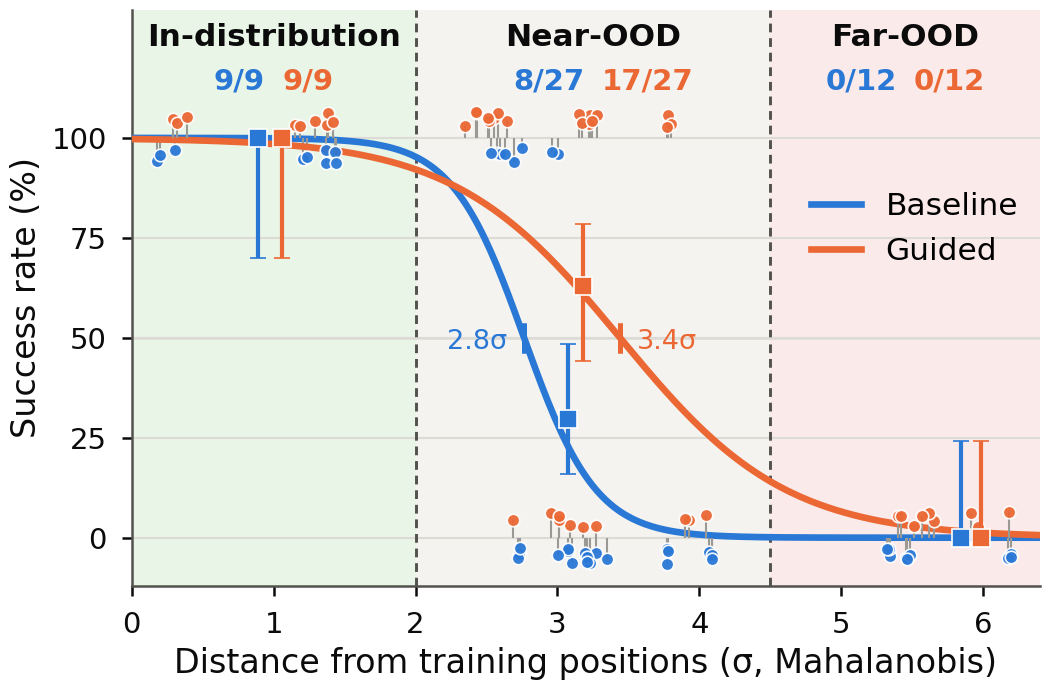}
    \caption{Positional generalization. Every rollout is plotted at the
    Mahalanobis distance of its measured cable position (dots near $100$ are
    successes, near $0$ failures), with per-zone success rates and $95\%$ Wilson
    intervals (squares) and a per-method logistic fit, crossing $50\%$ at
    $2.8\sigma$ (base) and $3.4\sigma$ (\ourmethod).}
    \label{fig:real_positional}
\end{figure}

In the near-OOD band the guidance raises the success rate by $+33.3$ points
($p=0.028$, Table~\ref{tab:real_positional}). It neither helps nor hurts
in-distribution, and does not recover the far-OOD failures, for a pooled gain of
$+18.8$ points ($p=0.10$) over the sweep (Figure~\ref{fig:real_positional_grid}c,d).
The per-method logistic fit restates this on a continuous scale
(Figure~\ref{fig:real_positional}): guidance pushes the $50\%$ crossing out by
$0.6\sigma$ and flattens the decay (Appendix~\ref{sec:appendix-positional}).
Qualitatively, most failures, guided or not, occur in the grasping phase
(Section~\ref{sec:real_failures}); in the failed guided rollouts the approach
still brought the end-effector over the cable.

\begin{figure}[!h]
    \centering
    \includegraphics[width=0.64\columnwidth]{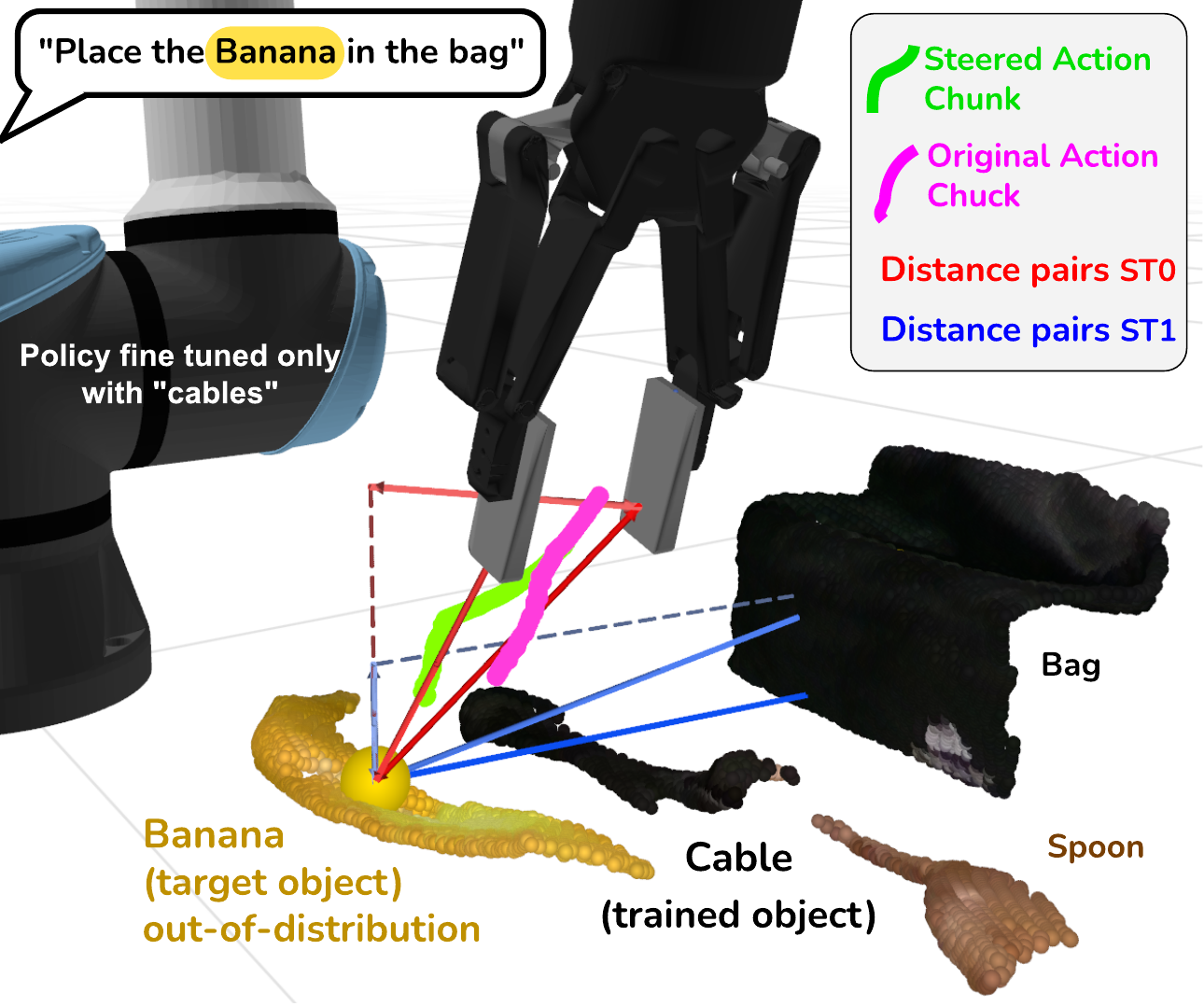}
    \caption{Disambiguating the target object in a cluttered scene. In red and blue, the distance pairs used per sub-trajectory (ST0 approach, ST1 transport).}
    \label{fig:real_disambiguation}

\end{figure}

\begin{figure}[!t]
    \centering
    \begingroup
    \captionsetup{type=table}
    \caption{Task variation. \emph{Named} is the rate at which the arm went for
    the object the prompt named; \emph{Task SR} also requires it to be
    bagged; \emph{Confusion} is the rate of grasping a distractor. The policy was only trained on the cable, the banana and the spoon are unseen.}
    \label{tab:real_objects}
    \adjustbox{max width=\columnwidth}{
    \begin{tabular}{lccccccc}
        \toprule
        & & \multicolumn{4}{c}{Named (\%)} & & \\
        \cmidrule(lr){3-6}
        Method & Trials & Cable & Banana & Spoon & All & Task SR (\%) & Confusion (\%) $\downarrow$ \\
        \midrule
        Base (no steering) & $24$ & 38 & 88 & 38 & 54.2 & 33.3 & 45.8 \\
        \ourmethod          & $24$ & \textbf{100} & \textbf{100} & \textbf{88} & \textbf{95.8} & \textbf{62.5} & \textbf{4.2} \\
        \bottomrule
    \end{tabular}
    }
    \endgroup

    \vspace{0.8em}
    \includegraphics[width=\columnwidth]{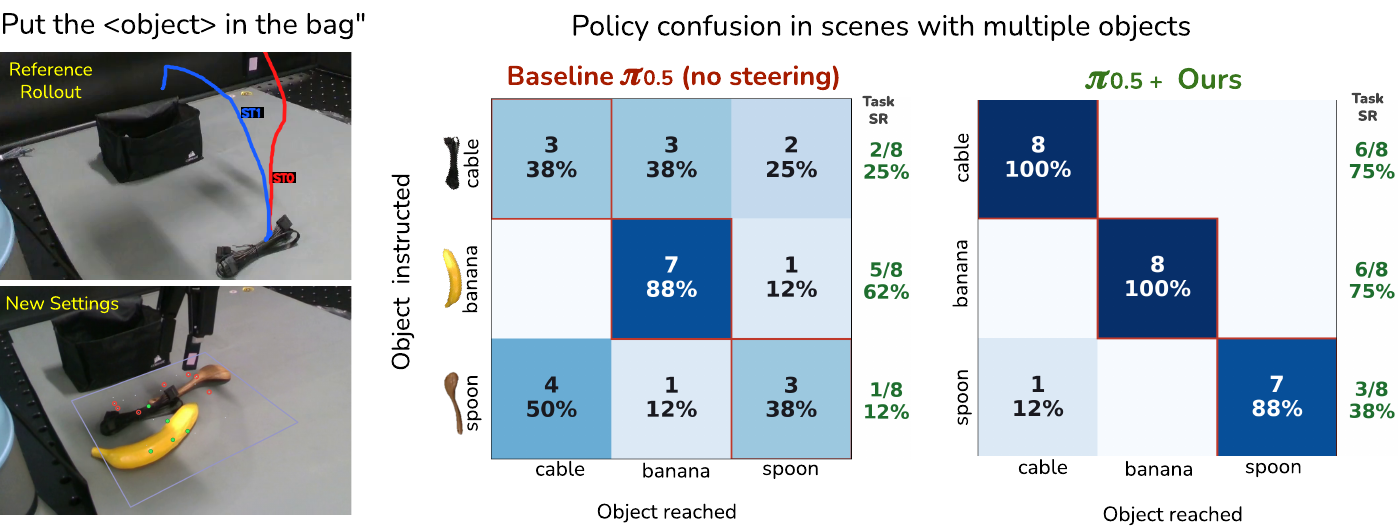}
    \caption{Task variation on cable bagging. Left: the single reference rollout
    above the new setting in which cable, banana and spoon share the workspace.
    Center, right: confusion matrices (instructed object in rows, reached object
    in columns) for the base and the guided policy.}
    \label{fig:real_tasks}
\end{figure}

\textbf{Q5: Can the policy be steered toward objects it was not trained on?} \label{sec:real_objects}%
The base policy was trained to bag a cable and nothing else, so can the same
frozen checkpoint be \emph{redirected} to a different object simply by changing
the relations the guidance tracks? We place the three objects of
Figure~\ref{fig:real_tasks} in the scene \emph{simultaneously} and name one in
the prompt. All three are elongated and share the graspable affordance the
policy overfit to, so the setting isolates its tendency to approach whichever
object its visual prior favours, regardless of the instruction.
Figure~\ref{fig:real_disambiguation} shows one such scene with the transferred
sub-trajectories and the relations anchored on the named object. A trial is
\emph{correct} when the arm grasps the named object and \emph{confused} when it
grasps a distractor.

The base policy is confused on $45.8\%$ of trials ($11/24$): prompted for the
spoon it goes for the cable in half the trials, and prompted for the cable it
splits evenly between cable and banana; only the banana, the most visually
distinctive, is reliably singled out ($88\%$). Under guidance, confusion falls to
$4.2\%$ ($1/24$) and the named object is reached on $95.8\%$ of trials. Task
success rises from $33.3\%$ to $62.5\%$, by less than the naming rate: steering
reliably decides \emph{which} object the frozen policy goes for, while bagging an
object it was never trained to handle remains limited by its low-level grasp.





\subsection{Failure modes}
\label{sec:real_failures}

\begin{figure}[!b]
    \centering
    \begin{subfigure}[b]{0.46\columnwidth}
        \centering
        \includegraphics[width=\linewidth]{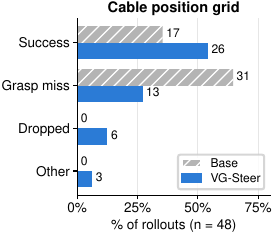}
        \caption{Positional variation.}
        \label{fig:real_failures_positional}
    \end{subfigure}\hfill
    \begin{subfigure}[b]{0.46\columnwidth}
        \centering
        \includegraphics[width=\linewidth]{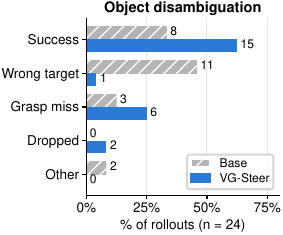}
        \caption{Task variation.}
        \label{fig:real_failures_disambiguation}
    \end{subfigure}
    \caption{Real-world outcome breakdown (share of rollouts, counts on bars).
    Residual failures concentrate in grasp and transport, which the guidance
    does not address.}
    \label{fig:real_failures}
\end{figure}

We annotate each real-world rollout with the first event that ended it:
\textbf{wrong target} (grasps an object the prompt did not name), \textbf{grasp
miss} (reaches the right object but never closes on it), \textbf{dropped},
and \textbf{other} (mostly perception dropouts and protective stops).

The two studies fail for opposite reasons under the base policy
(Figure~\ref{fig:real_failures}): in the positional sweep all $31$ failures are
grasp misses, the signature of reaching for a position never seen in training,
whereas in the task study $11$ of $16$ are wrong-target grasps. Guidance corrects
both, cutting grasp misses to $13$ and leaving a single wrong-target trial, and
the residual failures move downstream past contact, into drops during transport. The guidance acts on where the end-effector goes and which
object it goes to, not on contact, so grasping and transport become the limiting
stage once the approach is corrected, consistent with the far-OOD dead band of
Section~\ref{sec:real_positional}.


\section{Conclusion}
\label{sec:conclusion}

We presented \ourmethod, a visual-geometric policy steering framework that improves the OOD generalization of frozen flow-matching policies without any weight update. By retrieving task-relevant in-distribution demonstrations and abstracting them into a single guidance signal of object-relational distance pairs, tracked per timestep along VLM-selected axes, it injects differentiable corrections over position, orientation, and grasp directly into the policy's denoising process. On LIBERO-Pro, it outperforms VLS under pose and task-goal shifts and transfers to a second base policy without retuning, and on a physical UR5 (Section~\ref{sec:real_world}), under real open-vocabulary perception, it improves over the unsteered base under positional shift and redirects a single-task policy to unseen objects while nearly eliminating object confusion.

\textbf{Limitations.} \emph{(i) Perception.} The guidance is only as reliable as the grounding it is built on, which requires detecting, labeling, and localizing the task-relevant objects in both the demonstration and the live scene. Our simulation experiments use ground-truth segmentation and labels to isolate the steering signal; under real open-vocabulary perception (Appendix~\ref{sec:appendix-realperception}), missed detections, mislabeled instances, and imprecise masks or depth corrupt the distance pairs and substantially degrade performance. This gap in Table~\ref{tab:libero_full_breakdown} makes perception the dominant source of error. \emph{(ii) Sub-trajectory switching.} Progress depends on the bidirectional, sub-goal-gated phase transitions (Section~\ref{sec:guidance_injection}): when a sub-goal is mis-detected, the active relation set can advance too early, stall, or oscillate, steering toward the wrong relations. \emph{(iii) Balancing guidance and policy behavior.} The single global guidance scale, tempered by the line search and do-no-harm guard, mediates but does not resolve the tension between following the guidance and preserving the policy's learned behavior. In contact-rich interactions such as grasping or insertion, over-steering disrupts the policy's low-level motor skills while under-steering fails to correct the OOD error; an adaptive, interaction-aware schedule could improve this balance.


\vspace{-3mm}


\section*{ACKNOWLEDGMENT}
This work was in part supported by the Institute of Information \& Communications
Technology Planning \& Evaluation(IITP) grant funded by the Korea
government(MSIT) (No.RS-2025-25463654). This research was supported in part by Fujitsu.
\bibliographystyle{IEEEtran}
\bibliography{references}

\appendix
\section{Appendix}
\label{sec:appendix}

This appendix collects the implementation-specific details of the pipeline summarized in Sections~\ref{sec:method}--\ref{sec:implementation-details}. The values reported below are those used for the experiments unless stated otherwise.

\subsection{Perception pipeline}
\label{sec:appendix-perception}
Object instance masks are produced by an open-vocabulary, grounded pipeline rather than a single segmentation model. For each task-relevant object named in the instruction, a VLM (Gemini~3~Flash \cite{gemini}) proposes one grounded bounding box; SAM2 \cite{sam2} converts each box into a precise mask; and the masks are size- and erosion-filtered before centroids and surface points are recovered by unprojecting the depth image and fusing the available camera views. After the first frame, masks are propagated across frames with the SAM2 video tracker so that per-step perception is inexpensive; full VLM--SAM2 detection is rerun only when re-grounding is needed. The same pipeline is applied to both the demonstration frame and the live OOD frame, giving consistent, language-grounded labels across scenes. Segmentation therefore comes from grounded VLM boxes refined by SAM2, rather than from a single concept-level segmentation model.

\subsection{Demonstration analysis and VLM prompting}
\label{sec:appendix-demo-analysis}
\begin{figure}[t]
    \centering
    \includegraphics[width=\linewidth]{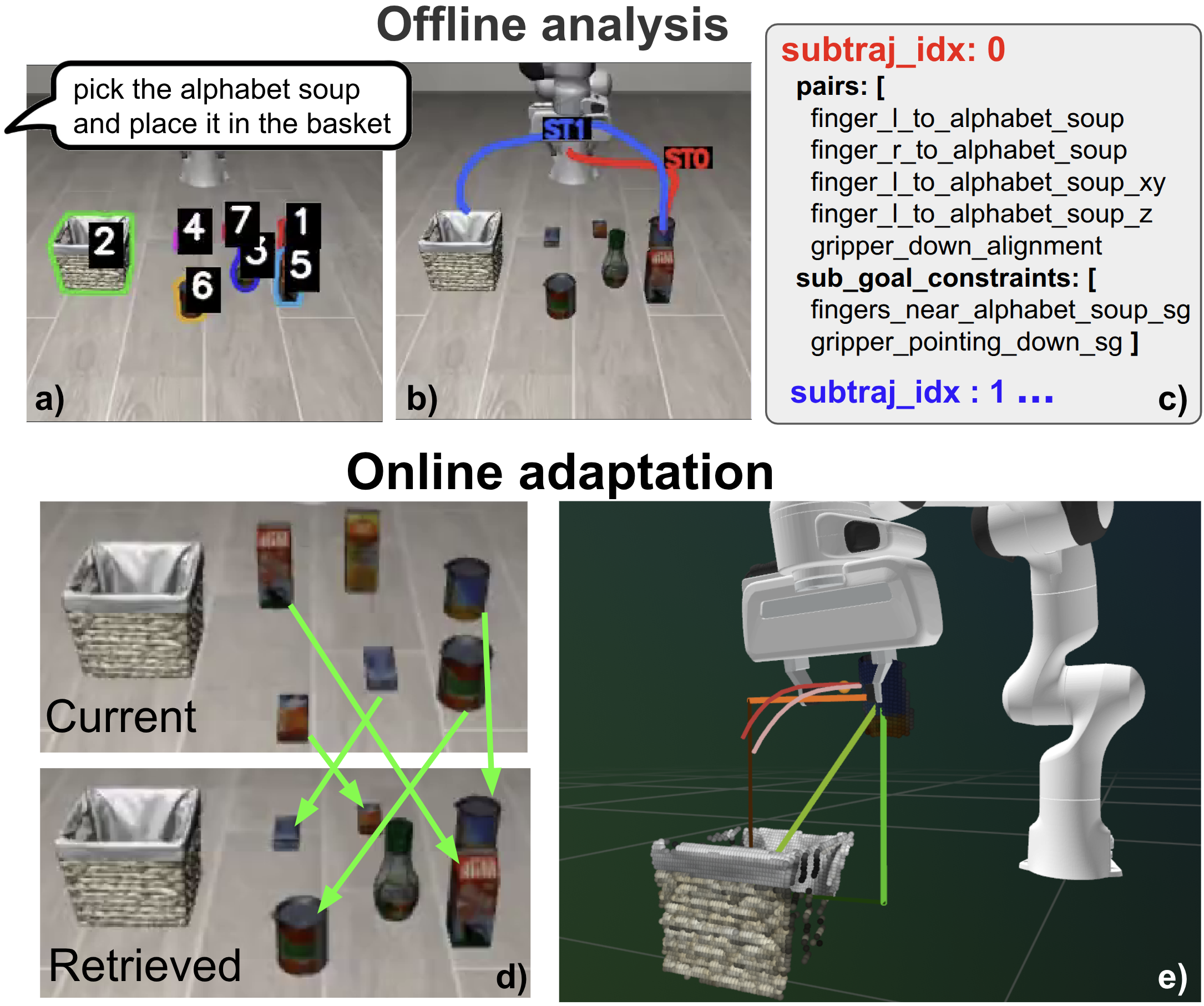}
    \caption{Visual-geometric guidance via VLM-selected distance pairs. \textbf{Offline} (a--c): the demonstration is segmented (Gemini~3~Flash + SAM2) and split into sub-trajectory phases (STRAP); the VLM selects each phase's entity pairs, projections, and sub-goal constraints. \textbf{Online} (d--e): the live scene is matched to the demonstration, and the guidance steers the frozen policy so the selected projected distances track the reference.}
    \label{fig:vg_guidance}
\end{figure}

The entity pairs, relation types, and projections of Section~\ref{sec:vg_guidance} are selected by a separate reasoning VLM (GPT-5.4) queried in a single-turn prompt with the task description and the two annotated images (segmentation overlay and STRAP sub-trajectory overlay). An optional iterative variant (up to three rounds, selecting pairs by alignment consistency across demonstrations) is available but disabled in the reported runs. VLM outputs and live object detections are recomputed per episode (caching disabled) so that results reflect the full online pipeline. Figure~\ref{fig:vg_guidance} walks through the offline abstraction and its online application.

\subsection{Sub-trajectory segmentation}
\label{sec:appendix-segmentation}
Following the velocity-based heuristic of STRAP \cite{strap}, the demonstration is split at low end-effector speed points: we threshold the end-effector speed (computed from the end-effector position trajectory) at $0.005$ and enforce a minimum sub-trajectory length of $20$ steps to avoid spurious fragmentation.

\subsection{Object-relational relations and per-relation weights}
\label{sec:appendix-relations}
Each relation specifies a \emph{source} and a \emph{distance type}. Sources are the end-effector, the left or right fingertip (obtained by forward kinematics from the end-effector pose and gripper aperture), an object rigidly held by the gripper (placed at a body-frame offset from the end-effector), or, for orientation and gripper relations, the end-effector pose and gripper channel directly. Distance types are: \textbf{centroid} (Euclidean), \textbf{surface} (minimum distance from the source to a sub-sampled object point cloud), \textbf{orientation} (signed cosine of a rotated body-frame vector $R\hat{e}$ with an axis, or its in-plane magnitude w.r.t.\ a plane), and \textbf{gripper} (metric aperture). The gripper channel is mapped from the policy's normalized action range to metric finger separation so that finger-kinematics and aperture relations are all expressed in metres.

The per-relation weight $w_n$ of Eq.~\ref{eq:guidance_loss} combines the VLM-assigned importance of relation $n$ with a \emph{relevance gate}: a relation is linearly ramped from inactive to fully active as the end-effector distance to the relevant object decreases from a far threshold ($0.15$\,m) to a near threshold ($0.05$\,m), so that a relation only steers the policy once the corresponding object is being approached. The reference distance trajectory $d^{*n}_{t'}$ is taken from the retrieved demonstration; when several demonstrations are available, their dynamic-time-warping-aligned spread yields a per-relation, per-step standard deviation that is used for alignment and diagnostics (it does not reweight the loss in the reported configuration).

\subsection{Trajectory alignment}
\label{sec:appendix-alignment}
Alignment runs online in the visual-geometric feature space. In the MIXED mode we use, a shared cursor is first matched to the reference by minimizing the mean squared error over the relevant relations, and then each relation is independently re-aligned within a window of half-width $5$ frames around that cursor. The cursor is \emph{not} monotonic: within the active sub-trajectory it may move forward or backward, so the guidance can track retries; per step it advances by at most $1.2$ action chunks to avoid large jumps. Distances are normalized per sub-trajectory so matching is driven by relative progress, and the optional temporal-pace term is disabled. Sub-trajectory switching is bidirectional and gated on sub-goal satisfaction, with a persistence of $10$ steps to reject transient signals. A softmax over each relation's alignment lag (temperature $5.0$, floor $0.05$) sets the per-relation temporal weighting that enters $w_n$.

\subsection{Action roll-out (dynamics model)}
\label{sec:appendix-rollout}
The dynamics model $\mathcal{F}$ of Section~\ref{sec:vg_guidance} is a kinematic integrator of the OSC velocity actions. From the current end-effector position $p_0$ and axis-angle orientation $r_0$ (with $R_0 = \exp r_0$), it integrates the chunk's per-step linear and angular velocities $(v_k, \omega_k)$ with an Euler (cumulative-sum) scheme,
\begin{equation}
\begin{aligned}
    &p_{t'} = p_0 + \sum_{k\le t'} v_k\,\Delta t, \qquad
    r_{t'} = r_0 + \sum_{k\le t'} \omega_k\,\Delta t, \\
    &R_{t'} = \exp(r_{t'}),
\end{aligned}
\end{equation}
where $\Delta t$ is the per-step duration. The gripper channel is absolute and read directly, mapped to a metric aperture as in Appendix~\ref{sec:appendix-relations}. Because $\mathcal{F}$ is differentiable, the guidance gradient flows through it back to the action chunk; any other differentiable dynamics model can be substituted without changing the rest of the pipeline. In the real world the policy predicts joint positions, so $\mathcal{F}$ is the analytic UR5 forward kinematics.

\subsection{Guidance injection and line search}
\label{sec:appendix-injection}
\begin{algorithm}[h]
\caption{Per-step golden-section line search for the guidance scale $\alpha^\star$.}
\label{alg:line_search}
\begin{algorithmic}[1]
\Require chunk $A^{(k)}_t$, unit direction $\hat{d}$, loss $\mathcal{L}_{guide}$, bound $\alpha_{\max}$, iterations $n$
\State define $f(\alpha) \equiv \mathcal{L}_{guide}(A^{(k)}_t + \alpha\,\hat{d})$,\quad $\varphi \gets (\sqrt{5}+1)/2$
\State $a \gets 0,\; b \gets \alpha_{\max}$
\While{$f(2b) < f(b)$} \Comment{expand bracket (capped at 5 doublings)}
    \State $b \gets 2b$
\EndWhile
\For{$n$ iterations} \Comment{golden-section narrowing}
    \State $u \gets b-(b-a)/\varphi,\quad w \gets a+(b-a)/\varphi$
    \If{$f(u) < f(w)$}
        \State $b \gets w$
    \Else
        \State $a \gets u$
    \EndIf
\EndFor
\State $\alpha \gets (a+b)/2$
\State \Return $\alpha$ \textbf{if} $f(\alpha) < f(0)$ \textbf{else} $0$ \Comment{do-no-harm}
\end{algorithmic}
\end{algorithm}

Guidance is applied at every replanning call and every denoising step. The gradient is evaluated at the policy's tentative next iterate $A^{(k)}_t + \Delta t\,v$ over the first predicted action chunk, which holds the executed actions. Our $\pi_{0.5}$ checkpoint predicts two consecutive $10$-step chunks per call; since it replans every $5$ steps, the second, look-ahead chunk is never executed and is left unguided.

\textbf{Gradient normalization.} By default the gradient is $\ell_2$-normalized to a unit direction, $\hat{d} = -g/\lVert g\rVert$, so the golden-section step size, not the raw gradient scale, sets the correction magnitude. This is the \emph{gradient normalization} component of the ablation (Table~\ref{tab:ablation}). Its ablated variants use the raw gradient $-g$ instead: the fixed-step row applies it with $c = 1$ and no line search, and the line-search row searches its step over $[0, 100]$ rather than $[0, 1]$, since the raw gradient is an order of magnitude or more smaller than the unit direction. The golden-section search (Algorithm~\ref{alg:line_search}) uses $20$ narrowing iterations and expands its bracket by at most five doublings; the do-no-harm guard returns a zero step whenever no positive step reduces the loss.

An optional multi-step variant decomposes the correction across relations or chunk timesteps: \emph{sequential} (greedy coordinate descent over relations, line-searched against the total loss), \emph{sum}/\emph{weighted} (independent per-relation steps, optionally scaled by $w_n$), and \emph{per-timestep} (the timestep analog of sequential). These are disabled in the reported runs, which use a single shared descent direction.

\subsection{Hyperparameters and base policies}
\label{sec:appendix-hyperparams}
The global guidance scale is $c = 1.0$, and the relevance thresholds are $(0.05, 0.15)$\,m across all benchmarks and base policies; only the line-search bound $\alpha_{\max}$ is rescaled per policy to match its action parameterization (velocity field versus score). Guidance is recomputed every $5$ environment steps (the policy replanning interval). We evaluate $\pi_{0.5}$ (OpenPI checkpoint \texttt{pi05\_libero\_grounded}; $224{\times}224$ inputs) and GR00T~N1.7 (\texttt{GR00T-N1.7-LIBERO}; $256{\times}256$ inputs), keeping each policy frozen. On LIBERO-Pro we run $10$ trials per task across the object, spatial, language, and goal/task perturbation conditions. The retrieval encoders are DINOv2 (\texttt{dinov2-base}) for vision and CLIP (\texttt{clip-vit-base-patch16}) for language.

\subsection{Full results breakdown}
\label{sec:appendix-breakdown}
Table~\ref{tab:libero_full_breakdown} reports the per-perturbation-type success rates behind the task-and-positional summary in the main text, including the near-saturated object and language perturbations. The base policies already reach $75$--$95\%$ on object and language shifts, so steering has little headroom there and can slightly perturb an already-correct trajectory; the gains concentrate on the task and positional shifts.

\begin{table}[t]
    \centering
    \caption{Per-perturbation-type success rate (\%) on LIBERO-Pro, averaged over the four suites. The base is near-saturated on object and language, so the overall average understates the task/positional gains. ``GT labels'' / ``Gemini 3 Flash'' denote the perception used for \ourmethod.}
    \label{tab:libero_full_breakdown}
    \adjustbox{max width=\columnwidth}{
    \begin{tabular}{lccccc}
        \toprule
        Method & Task & Pos & Object & Lang. & Overall \\
        \midrule
        \multicolumn{6}{l}{\textit{$\pi_{0.5}$}} \\
        \quad No steering            & 26.3 & 26.3 & 83.0 & 92.3 & 56.9 \\
        \quad VLS (reproduced, full) & 32.0 & 27.6 & 45.8 & 61.5 & 41.7 \\
        \quad VLS (grad-only)        & 17.3 & 21.5 & 18.3 & 26.8 & 20.9 \\
        \quad \ourmethod (GT labels)       & 40.8 & 34.3 & 65.8 & 75.0 & 53.9 \\
        \quad \ourmethod (Gemini 3 Flash)  & 24.8 & 24.5 & 47.5 & 55.3 & 38.0 \\
        \addlinespace
        \multicolumn{6}{l}{\textit{GR00T N1.7}} \\
        \quad No steering            & 19.3 & 0.5  & 75.5 & 90.5 & 46.4 \\
        \quad \ourmethod (GT labels)       & 20.0 & 3.5  & 60.8 & 75.8 & 40.0 \\
        \quad \ourmethod (Gemini 3 Flash)  & 14.5 & 3.5  & 44.0 & 57.0 & 29.8 \\
        \bottomrule
    \end{tabular}
    }
\end{table}

\subsection{Real open-vocabulary perception}
\label{sec:appendix-realperception}
Under real open-vocabulary perception (grounded VLM detection followed by SAM2 masks, with no simulator labels, unlike the ground-truth grounding used in the main results), $\pi_{0.5}$+\ourmethod reaches $24.8$ / $24.5$ on the task / positional perturbations (Table~\ref{tab:libero_full_breakdown}), well below the $40.8$ / $34.3$ obtained with ground-truth grounding. This difference quantifies how much current performance is bottlenecked by perception rather than by the steering itself. Within real perception, the choice of grounding backbone matters: Table~\ref{tab:perception_ablation} compares Gemini Robotics-ER~1.6 and Gemini~3~Flash, with Gemini~3~Flash performing best ($24.7$ vs.\ $21.1$ averaged over the task and positional suites); we use it throughout.

\begin{table}[t]
    \centering
    \caption{Open-vocabulary grounding-backbone ablation for $\pi_{0.5}$+\ourmethod under real perception (success rate \%; ``Pos'' denotes the positional/swap perturbation). Gemini~3~Flash outperforms Gemini Robotics-ER~1.6 on the task and positional suites.}
    \label{tab:perception_ablation}
    \begin{tabular}{llcc}
        \toprule
        Suite & Pert. & ER-1.6 & Gemini 3 Flash \\
        \midrule
        Goal       & Task & 9.0  & 11.0 \\
        Goal       & Pos  & 23.0 & 25.0 \\
        Spatial    & Task & 40.0 & 49.0 \\
        Spatial    & Pos  & 33.0 & 32.0 \\
        libero\_10 & Task & 5.0  & 6.0  \\
        libero\_10 & Pos  & 9.0  & 8.0  \\
        Object     & Task & 23.0 & 33.0 \\
        Object     & Pos  & 27.0 & 33.0 \\
        \midrule
        \multicolumn{2}{l}{Avg (task+pos)} & 21.1 & \textbf{24.7} \\
        \bottomrule
    \end{tabular}
\end{table}

\subsection{Real-world platform details}
\label{sec:appendix-realworld}
\textbf{Robot and control.} A UR5 6-DoF arm with a Robotiq~2F-140 parallel gripper ($0$--$140$\,mm stroke). The arm is
commanded in joint space + gripper,  matching the action space the base policies emit; the policy emits 50-step chunks replanned
every 25 steps. The gripper aperture is mapped from the policy's normalized channel to metric finger separation as in
Appendix~\ref{sec:appendix-relations}.

\textbf{Cameras and calibration.} All object detection and 3D reconstruction for the guidance are performed from a
\emph{single fixed external} Intel RealSense~D435, mounted on the right side of the arm. Intrinsics are taken from the camera's factory
calibration. Extrinsics are obtained \emph{markerlessly} with Kalib \cite{kalib}, which tracks a reference point on the
robot with a visual foundation model and recovers the camera-to-robot transform with a PnP solve from forward
kinematics; no fiducial markers, mesh models or network training are required, and a full calibration takes seconds.
This matters in practice because the guidance consumes metric object centroids and surface points unprojected from
depth: a calibration error translates directly into a bias on every distance pair. Because only one
view is available, the multi-view fusion described in Appendix~\ref{sec:appendix-perception} reduces to a single view,
so object centroids are recovered from the visible surface alone and self-occluded geometry is not observed.

\textbf{Base policy and reference rollout.} The base policy is $\pi_{0.5}$ \cite{pi05} (base), kept frozen. The
demonstration database holds a single successful rollout of that same policy, recorded in the nominal configuration
with the cable (Section~\ref{sec:real_world}). Guidance hyperparameters are unchanged from Appendix~\ref{sec:appendix-hyperparams}.

\subsection{Real-world perception timing}
\label{sec:appendix-realtiming}
The simulation profiling of Figure~\ref{fig:profiling} assumes privileged object states. On hardware, the guidance also
pays for real open-vocabulary perception (Appendix~\ref{sec:appendix-perception}). We report it in two parts: a
once-per-episode first-frame setup, and the per-replan cost inside the inference loop.

\textbf{First-frame setup.} Table~\ref{tab:realtiming_setup} breaks down the setup over $6$ runs on the same scene with
a fresh VLM cache each run. The six VLM round-trips sum to $20.6\pm3.1$\,s, so roughly $70\%$ of the $29.5\pm3.1$\,s
total is API latency; live-object inference is the noisiest call. Process startup before the episode is separate
(retrieval encoders $1.6\pm0.6$\,s, perception clients $3.4\pm0.4$\,s).

\begin{table}[h]
    \centering
    \small
    \caption{First-frame setup, mean $\pm$ sd over $6$ runs.}
    \label{tab:realtiming_setup}
    \begin{tabular}{lr}
        \toprule
        Stage & Wall time \\
        \midrule
        Retrieval (DINOv2 + CLIP embed and match) & $0.43\pm0.01$\,s \\
        Demo seed frame: object names (VLM)       & $2.1\pm0.3$\,s \\
        Demo seed frame: box detection (VLM)      & $2.1\pm0.4$\,s \\
        Demo seed frame: SAM2 masks               & $0.16\pm0.14$\,s \\
        Pair and constraint selection (VLM)       & $9.4\pm0.6$\,s \\
        Demo analysis total                       & $14.9\pm0.8$\,s \\
        Reference distance trajectory ($57$ frames) & $3.2\pm0.4$\,s \\
        Live-object inference (VLM)               & $3.0\pm1.7$\,s \\
        Live seed-frame detection (boxes + SAM2)  & $2.6\pm0.5$\,s \\
        Equivalence mapping (VLM)                 & $3.1\pm0.4$\,s \\
        \midrule
        \textbf{Total}                            & $\mathbf{29.5\pm3.1}$\,\textbf{s} \\
        \bottomrule
    \end{tabular}
\end{table}

\textbf{Inference loop.} We replayed detect-then-track over the primary view of six recorded guided rollouts at the
$25$-step replan cadence, using each rollout's own object names; every object was found on every frame
(Table~\ref{tab:realtiming_loop}). Full detection runs once per episode; afterwards the loop pays only mask
propagation and back-projection, $113\pm2$\,ms per replan.

\begin{table}[h]
    \centering
    \small
    \caption{Per-replan perception cost in the inference loop.}
    \label{tab:realtiming_loop}
    \begin{tabular}{lr}
        \toprule
        Measurement & Value \\
        \midrule
        Detection, once per episode {{\tiny(VLM boxes + SAM2)}} & $2.6\pm0.6$\,s \\
        Track call per replan (replay, $63$ frames)    & $46\pm4$\,ms \\
        Tracker step alone (replay)                    & $37.5\pm3.5$\,ms \\
        Tracker step alone (in the loop, $80$ rollouts) & $48\pm14$\,ms \\
        \midrule
        \textbf{Perception block in the live loop} & $\mathbf{113\pm2}$\,\textbf{ms} \\
        \bottomrule
    \end{tabular}
\end{table}

\subsection{Positional OOD analysis}
\label{sec:appendix-positional}
\textbf{Training distribution.} The $28$ training demonstrations of ``Put all cables in the bag.'' yield $110$ cable
positions, one per gripper closure near the table, read from the demonstrations' proprioception in the robot base
frame; these are grasp points and therefore a proxy for where the cable lay. On the table they form a weakly
correlated Gaussian ($\rho=-0.07$) centered in cell B2 of the $4\times4$ evaluation grid, with a standard deviation
of $6.2$\,cm across columns and $5.6$\,cm across rows. Each experiment rollout's cable position, measured from its
first frame with the perception pipeline of Appendix~\ref{sec:appendix-perception}, is expressed as a Mahalanobis
distance from that Gaussian, in standard deviations. The two sets of positions therefore reach us through different
channels, demonstration proprioception at gripper closure against perception on the first frame, so the distances
carry the offset between them. A systematic offset would shift all reported distances by a common amount; it would
not change the ordering of the cells or their zone assignment, which follow from the grid geometry.

\textbf{Statistics.} We analyze success against that distance in two ways: binned into three zones, with $95\%$
Wilson score intervals per method and a two-sided Fisher exact test between methods per zone
(Table~\ref{tab:real_positional}), and as a logistic regression of success on distance fitted per method, whose
$50\%$ crossing we report (Table~\ref{tab:positional_logistic}). Intervals on the difference between methods
use Newcombe's hybrid score method: $[+7, +54]$ points in the near-OOD band, $[+2, +41]$ pooled over OOD and
$[-1, +37]$ over the full sweep. Fisher's exact test is conservative on small, discrete samples, which is why the
pooled OOD comparison gives $p=0.051$ while the corresponding interval excludes zero. Both procedures also treat the
$48$ rollouts per method as independent, which they are not: each placement is run by both methods, a paired design
these unpaired tests ignore, and the three trials in a cell share a nominal placement. We therefore report all of
these as descriptive summaries rather than as inference over independent trials.

\textbf{Zones.} We assign each rollout to a zone from its own measured position rather than from its nominal cell
(Table~\ref{tab:positional_zones}). Both boundaries lie in gaps of the distance distribution: no rollout falls
between $1.4$ and $2.3\sigma$ or between $4.1$ and $5.3\sigma$, so any cut placed inside a gap yields the same
assignment, and the two rollouts of a placement are assigned together unless their measured positions differ by more
than the $0.9\sigma$ gap width. Table~\ref{tab:positional_cells} lists the distance of every cell center, and
Figure~\ref{fig:real_positional_grid} shows the per-cell outcomes projected into the camera view.

\begin{table}[t]
    \centering
    \caption{Zone definition for the positional sweep, by Mahalanobis distance of the measured cable position from
    the training positions.}
    \label{tab:positional_zones}
    \adjustbox{max width=\columnwidth}{
    \begin{tabular}{llll}
        \toprule
        Zone & Distance & Cells & Rollouts per method \\
        \midrule
        In-distribution & below $2.0\sigma$        & B1, B2, B3              & $9$ \\
        Near-OOD        & $2.0$ to $4.5\sigma$     & A1 to A4, B4, C1 to C4  & $27$ \\
        Far-OOD         & above $4.5\sigma$        & D1 to D4                & $12$ \\
        \bottomrule
    \end{tabular}
    }
\end{table}

\begin{table}[t]
    \centering
    \caption{Cell centers of the $4\times4$ grid in the training distribution's standardized coordinates and as a
    Mahalanobis distance, sorted by distance. The distance uses the full covariance, which is weakly correlated, so
    it is not the Euclidean norm of the two $z$ columns. Centers are listed for reference only; zone assignment uses
    each rollout's own measured position.}
    \label{tab:positional_cells}
    \adjustbox{max width=0.9\columnwidth}{
    \begin{tabular}{lcccl}
        \toprule
        Cell & $z$ columns & $z$ rows & Mahalanobis & Zone \\
        \midrule
        B2       & $-0.13$ & $-0.26$ & $0.30$ & in-distribution \\
        B3       & $1.09$  & $-0.26$ & $1.10$ & in-distribution \\
        B1       & $-1.35$ & $-0.26$ & $1.39$ & in-distribution \\
        B4       & $2.30$  & $-0.26$ & $2.30$ & near-OOD \\
        C2       & $-0.13$ & $2.44$  & $2.44$ & near-OOD \\
        C1       & $-1.35$ & $2.44$  & $2.71$ & near-OOD \\
        C3       & $1.09$  & $2.44$  & $2.74$ & near-OOD \\
        A2       & $-0.13$ & $-2.95$ & $2.97$ & near-OOD \\
        A3       & $1.09$  & $-2.95$ & $3.08$ & near-OOD \\
        A1       & $-1.35$ & $-2.95$ & $3.33$ & near-OOD \\
        C4       & $2.30$  & $2.44$  & $3.47$ & near-OOD \\
        A4       & $2.30$  & $-2.95$ & $3.62$ & near-OOD \\
        D1 to D4 & $-1.35$ to $2.30$ & $5.13$ & $5.1$ to $5.8$ & far-OOD \\
        \bottomrule
    \end{tabular}
    }
\end{table}

\begin{table}[t]
    \centering
    \caption{Logistic fit of success against Mahalanobis distance, per method. The $50\%$ crossing is annotated in
    Figure~\ref{fig:real_positional}; the slope is the logit decrease per standard deviation. Both fits are
    constrained almost entirely by the near-OOD trials, since the other two zones are saturated for both methods,
    and neither is supported by data beyond $4.5\sigma$.}
    \label{tab:positional_logistic}
    \begin{tabular}{lcc}
        \toprule
        Logistic fit & Base (no steering) & \ourmethod \\
        \midrule
        $50\%$ crossing & $2.8\sigma$ & $3.4\sigma$ \\
        Slope           & $3.9$ per $\sigma$ & $1.7$ per $\sigma$ \\
        \bottomrule
    \end{tabular}
\end{table}

\subsection{VLS reproduction}
\label{sec:appendix-vls}
The VLS paper reports $38.5$ / $35.1$ ($36.8$ average) on the task / positional perturbations; our reproduced full run reaches $32.0$ / $27.6$. The configuration matches the paper's hyperparameters (batch $20$, $50$ denoising steps, $200$ episodes per suite, guidance scale, and resampling/diversity settings) and the same VLMs (GPT-5.1 and Gemini~3~Flash). As in the released LIBERO evaluation code, object segmentation and labels come from the simulator, the same input \ourmethod receives in Table~\ref{tab:main_results}. The residual gap is attributable to (i)~a different, re-exported base checkpoint (\texttt{pi05\_libero\_finetuned\_v044}); (ii)~routing all VLM calls through OpenRouter rather than the native OpenAI/Google APIs; and (iii)~single-run stochasticity (guidance code sampled at temperature $1.0$, live stage decisions, unpinned seeds) together with provider model drift since the paper's runs.

The \emph{grad-only} VLS row instead runs the public repository defaults, which set \texttt{sample\_batch\_size}$=1$. With a single particle, VLS's Feynman--Kac resampling and RBF diversity initialization are both disabled, so the method degenerates to pure gradient steering (the paper's ``w/o FKD, w/o RBF'' ablation), and it uses $10$ denoising steps. Because it receives the same simulator-provided segmentation and labels and has a comparable compute budget, this row isolates the gradient-steering signal and is the appropriate point of comparison for our single-chain gradient guidance.


\onecolumn
\subsection{Worked example: offline demonstration analysis}
\label{sec:appendix-offline-example}
\begin{figure}[h!]
    \centering
    \begin{subfigure}[t]{0.30\linewidth}
        \centering
        \includegraphics[width=\linewidth]{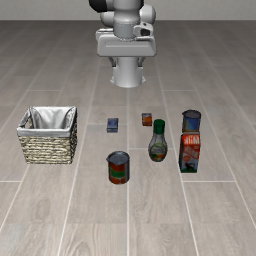}
        \caption{Original RGB scene.}
        \label{fig:offline-example-rgb}
    \end{subfigure}
    \hfill
    \begin{subfigure}[t]{0.30\linewidth}
        \centering
        \includegraphics[width=\linewidth]{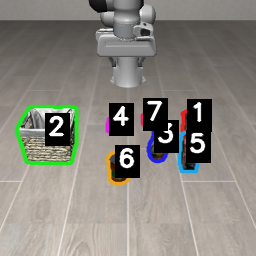}
        \caption{Segmentation overlay with numeric instance IDs.}
        \label{fig:offline-example-seg}
    \end{subfigure}
    \hfill
    \begin{subfigure}[t]{0.30\linewidth}
        \centering
        \includegraphics[width=\linewidth]{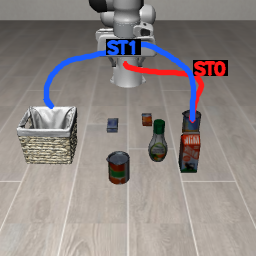}
        \caption{STRAP sub-trajectory overlay (ST0 approach, ST1 transport).}
        \label{fig:offline-example-subtraj}
    \end{subfigure}
    \caption{The three images provided to the reasoning VLM for the worked example of Appendix~\ref{sec:appendix-offline-example}: (a)~the raw RGB observation, (b)~the segmentation overlay that labels every object with its instance ID, and (c)~the end-effector path split into color-coded numbered sub-trajectories.}
    \label{fig:offline-example-inputs}
\end{figure}

This subsection reproduces a complete offline demonstration-analysis query end to end, for the task \emph{``pick up the alphabet soup and place it in the basket''}. The reasoning VLM (Section~\ref{sec:appendix-demo-analysis}) is given the system prompt below, together with the user message: the task description, the object list, the STRAP sub-trajectory split, and the three annotated images of Figure~\ref{fig:offline-example-inputs}. It returns the structured set of entity pairs, projections, and per-sub-trajectory sub-goal constraints shown in the assistant response, which the pipeline then compiles into the visual-geometric relations used for steering. The full message blocks are reproduced verbatim below.

\paragraph{System message.}
\begin{lstlisting}[style=transcript]
You are a robotics expert. Given a manipulation task description, a scene image (original RGB), a segmentation-overlay image that labels every object with its numerical instance ID, a sub-trajectory overlay image showing the robot end-effector path split into numbered sub-trajectories (each drawn in a distinct bright colour and labelled ST0, ST1, …), and a list of objects (segmentation ID + name), select the most informative pairs of entities whose pairwise distance serves as a useful heuristic for guiding task execution.

The special entity "ee" refers to the robot end-effector. The entities "gripper_l" and "gripper_r" refer to the left and right gripper finger tips respectively. Prefer "gripper_l" and "gripper_r" for any task that involves grasping, manipulating, or releasing objects with the fingers — they capture finger-level proximity (e.g. each finger to the target object) and gripper aperture ("gripper_l" to "gripper_r"). Only fall back to "ee" for tasks that do not involve the gripper fingers (e.g. pushing, pointing, or tool use where the end-effector as a whole is the relevant contact point).

For each pair choose the most appropriate distance type:
  - "centroid": Euclidean distance between the 3-D centroids of the two entities.
  - "surface": Shortest distance between the surfaces of the two entities (useful when contact/proximity matters).
  - "orientation": Projection of a body-frame vector of the end-effector onto a world-space reference direction. By default the tracked vector is the gripper's body z-axis (its "approach direction"), but you can set "ee_vector" to any body-frame direction (e.g. [1,0,0] for the gripper x-axis). Use when the orientation of the end-effector matters — e.g. ensuring the gripper approaches a target from above rather than from the side. Both "from_entity" and "to_entity" must be "ee" (the pair self-references). Requires either "axis" or "plane" (see below).
  - "gripper_aperture": Scalar gripper opening (metres). Both entities must be the gripper fingers — set "from_entity": "gripper_l" and "to_entity": "gripper_r".

Additionally, any pair can include at most ONE of the following optional component selectors to restrict the measurement to a subspace:
  - "axis" [x,y,z]: For positional pairs (centroid/surface), signed 1-D projection of the distance vector onto this direction — e.g. axis=[0,0,1] for vertical clearance during a lift, axis=[1,0,0] for progress along X during an approach. For "orientation" pairs, signed cosine of the angle between the tracked body vector and this direction (in [-1, 1]) — e.g. axis=[0,0,-1] gives +1 when the body vector points straight down ("approach from above") and -1 when it points straight up.
  - "plane" [nx,ny,nz]: For positional pairs, magnitude of the in-plane component after removing the contribution along this normal (2-D). Use for lateral alignment — e.g. plane=[0,0,1] gives XY-plane distance. For "orientation" pairs, magnitude of the tracked body vector projected into the plane perpendicular to the normal (in [0, 1]) — e.g. plane=[0,0,1] is 0 when the body vector is exactly vertical and 1 when it is fully horizontal.
  - "ee_vector" [x,y,z]: Only for "orientation" pairs. Specifies which body-frame direction of the end-effector to track. Defaults to [0,0,1] (z-axis, the "approach direction"). Set to [1,0,0] to track the gripper x-axis, [0,1,0] for the y-axis, or any arbitrary body-frame direction. Useful when the task cares about a sideways alignment rather than approach direction — e.g. pouring, where the gripper's x-axis must point away from the target container, or wiping, where the gripper's y-axis must remain aligned with the travel direction.

Set both "axis" and "plane" to null when the full 3-D distance is the right signal (positional pairs only — "orientation" pairs always need one).

Group your output by sub-trajectory. For each sub-trajectory phase, provide a single "reasoning" string (1–2 sentences) that describes the goal of that phase and why the selected pairs are informative for it. Use subtraj_idx null for pairs that are equally relevant across all sub-trajectories.

For each sub-trajectory you must also define sub-goal constraints:

SUB-GOAL CONSTRAINTS ("sub_goal"):
  Conditions that must be satisfied at the END of the sub-trajectory. They determine when to transition to the next sub-trajectory. Example: the end-effector must be within 3 cm of the mug before the grasp phase ends.

  IMPORTANT: Sub-goal constraints must depend ONLY on the CURRENT state of the objects and end-effector — they may compare positions of entities visible in the scene at the moment of evaluation, but must NOT reference past positions, initial positions, or any temporal history. If you need a height reference, use a currently visible entity (e.g. the table, a container, or another scene object), not the object's own past position.

Each constraint is expressed as a Python lambda string with the signature:
  lambda ee, keypoints, ee_R: <scalar expression>
where "ee" is a numpy array of shape (3,) with the end-effector XYZ position, "keypoints" is a dict mapping entity IDs (int or "ee") to numpy arrays of shape (3,) with their 3-D centroid positions, and "ee_R" is the world-from-EE rotation matrix as a numpy array of shape (3, 3). Within "ee_R", the third column "ee_R[:, 2]" is the gripper's body z-axis ("approach direction") expressed in world coordinates — use it to write orientation constraints such as "the gripper must point downward at end of approach". "ee_R[:, 0]" and "ee_R[:, 1]" are the other two body axes if a yaw-style constraint is needed. The constraint is SATISFIED when the returned scalar is <= 0. Use np for numpy operations. Entity IDs in keypoints match the segmentation IDs in the object list; "ee" maps to the end-effector centroid.

The lambda is evaluated in a restricted sandbox where only the "np" (numpy) module is available. Python built-ins such as max/min/abs/sum are NOT accessible — use numpy equivalents (np.maximum, np.minimum, np.abs, np.sum) instead. The lambda must return a scalar (a Python float or 0-D numpy value).

Example 1 — pick up a mug (2 sub-trajectories: approach ST0, lift ST1):
{
  "subtraj_idx": 0,
  "reasoning": "During the approach phase the fingers must close in on the mug, so the surface distances from each finger to the mug are the primary signals. The gripper must also approach from above (z-axis pointing down), maintain vertical clearance, and keep the fingers horizontally aligned with the handle.",
  "pairs": [
    {"from_entity": "gripper_l", "to_entity": 2,    "label": "finger_l_to_mug",       "distance_type": "surface",     "axis": null,        "plane": null},
    {"from_entity": "gripper_r", "to_entity": 2,    "label": "finger_r_to_mug",       "distance_type": "surface",     "axis": null,        "plane": null},
    {"from_entity": "gripper_l", "to_entity": 2,    "label": "finger_l_to_mug_z",     "distance_type": "centroid",    "axis": [0, 0, 1],   "plane": null},
    {"from_entity": "ee",        "to_entity": "ee", "label": "gripper_down_alignment", "distance_type": "orientation", "axis": [0, 0, -1], "plane": null, "ee_vector": null},
    {"from_entity": "ee",        "to_entity": "ee", "label": "gripper_side_alignment", "distance_type": "orientation", "axis": [0, 0, 1],  "plane": null, "ee_vector": [0, 1, 0]}
  ],
  "sub_goal_constraints": [
    {
      "label": "fingers_near_mug_sg",
      "description": "Both finger tips must be within 8 cm of the mug centroid at the end of approach.",
      "expression": "lambda ee, keypoints, ee_R: np.maximum(np.linalg.norm(keypoints['gripper_l'] - keypoints[2]), np.linalg.norm(keypoints['gripper_r'] - keypoints[2])) - 0.08",
      "reasoning": "Ensures the fingers have reached grasping distance before transitioning to the lift phase."
    },
    {
      "label": "gripper_pointing_down_sg",
      "description": "Gripper z-axis must be within ~30° of straight-down at the end of approach (cos(30°) ≈ 0.85).",
      "expression": "lambda ee, keypoints, ee_R: 0.85 - np.dot(ee_R[:, 2], np.array([0.0, 0.0, -1.0]))",
      "reasoning": "Confirms the gripper is well-aligned to a top-down grasp before the lift phase begins."
    }
  ]
},
{
  "subtraj_idx": 1,
  "reasoning": "The lift phase raises the grasped mug off the table; the Z-axis projection captures vertical progress while ignoring lateral offset.",
  "pairs": [
    {"from_entity": 2, "to_entity": 1, "label": "mug_table_z", "distance_type": "centroid", "axis": [0, 0, 1], "plane": null}
  ],
  "sub_goal_constraints": [
    {
      "label": "mug_lifted_sg",
      "description": "Mug centroid must be at least 10 cm above the table centroid.",
      "expression": "lambda ee, keypoints, ee_R: 0.10 - (keypoints[2][2] - keypoints[1][2])",
      "reasoning": "Confirms the mug has been successfully lifted before the episode ends."
    }
  ]
}

Example 2 — Pick the object and place it in the container (2 sub-trajectories: approach ST0, transport ST1):
{
  "subtraj_idx": 0,
  "reasoning": "ST0 is the approach-and-grasp phase for the object. The most informative signals are the finger-to-object surface distances and maintaining horizontal alignment between the fingers and the object.",
  "pairs": [
    {"from_entity": "gripper_l", "to_entity": 1,           "label": "finger_l_to_object",    "distance_type": "surface",  "axis": null, "plane": null},
    {"from_entity": "gripper_r", "to_entity": 1,           "label": "finger_r_to_object",    "distance_type": "surface",  "axis": null, "plane": null},
    {"from_entity": "gripper_l", "to_entity": 1,           "label": "finger_l_to_object_xy", "distance_type": "centroid", "axis": null, "plane": [0, 0, 1]}
  ],
  "sub_goal_constraints": [
    {
      "label": "fingers_near_object_sg",
      "description": "Both finger tips must be within 7 cm of the object centroid at the end of the approach phase.",
      "expression": "lambda ee, keypoints, ee_R: np.maximum(np.linalg.norm(keypoints['gripper_l'] - keypoints[1]), np.linalg.norm(keypoints['gripper_r'] - keypoints[1])) - 0.07",
      "reasoning": "This indicates the fingers have reached grasping proximity to the object before transitioning to transport."
    }
  ]
},
{
  "subtraj_idx": 1,
  "reasoning": "ST1 transports the grasped object and lowers it into the container. The key signals are horizontal alignment between object and container and then proximity to the container interior while maintaining enough height to avoid premature contact with the container rim.",
  "pairs": [
    {"from_entity": 1, "to_entity": 2, "label": "object_to_container_xy", "distance_type": "centroid", "axis": null,      "plane": [0, 0, 1]},
    {"from_entity": 1, "to_entity": 2, "label": "object_to_container",    "distance_type": "centroid", "axis": null,      "plane": null},
    {"from_entity": 1, "to_entity": 2, "label": "object_to_container_z",  "distance_type": "centroid", "axis": [0, 0, 1], "plane": null}
  ],
  "sub_goal_constraints": [
    {
      "label": "object_in_container_xy_sg",
      "description": "Object centroid must be horizontally within 6 cm of the container centroid at the end of placement.",
      "expression": "lambda ee, keypoints, ee_R: np.linalg.norm((keypoints[1] - keypoints[2])[:2]) - 0.06",
      "reasoning": "This ensures the object is laterally aligned with the container opening at the end of the placement motion."
    },
    {
      "label": "object_lowered_into_container_sg",
      "description": "Object centroid must be no more than 8 cm above the container centroid at the end of placement.",
      "expression": "lambda ee, keypoints, ee_R: (keypoints[1][2] - keypoints[2][2]) - 0.08",
      "reasoning": "This confirms the object has been lowered into the container region rather than remaining high above it."
    }
  ]
}

Example 3 — Grasp a small object and transport it to a target zone (3 sub-trajectories: approach ST0, grasp-and-lift ST1, transport ST2):
{
  "subtraj_idx": 0,
  "reasoning": "ST0 is the approach phase: the fingers must close in on the object with good horizontal alignment before grasping.",
  "pairs": [
    {"from_entity": "gripper_l", "to_entity": 1, "label": "finger_l_to_object",    "distance_type": "surface",  "axis": null, "plane": null},
    {"from_entity": "gripper_r", "to_entity": 1, "label": "finger_r_to_object",    "distance_type": "surface",  "axis": null, "plane": null},
    {"from_entity": "gripper_l", "to_entity": 1, "label": "finger_l_to_object_xy", "distance_type": "centroid", "axis": null, "plane": [0, 0, 1]}
  ],
  "sub_goal_constraints": [
    {
      "label": "fingers_near_object_sg",
      "description": "Both finger tips must be within 7 cm of the object centroid.",
      "expression": "lambda ee, keypoints, ee_R: np.maximum(np.linalg.norm(keypoints['gripper_l'] - keypoints[1]), np.linalg.norm(keypoints['gripper_r'] - keypoints[1])) - 0.07",
      "reasoning": "The fingers have reached grasping proximity."
    }
  ]
},
{
  "subtraj_idx": 1,
  "reasoning": "ST1 is the grasp-and-lift phase. The gripper fingers must close around the object and then lift it. Monitoring finger-to-object surface distances confirms the fingers are positioned around the object; the gripper aperture tracks finger closure.",
  "pairs": [
    {"from_entity": "gripper_l", "to_entity": 1,           "label": "finger_l_to_object",  "distance_type": "surface",  "axis": null, "plane": null},
    {"from_entity": "gripper_r", "to_entity": 1,           "label": "finger_r_to_object",  "distance_type": "surface",  "axis": null, "plane": null},
    {"from_entity": "gripper_l", "to_entity": "gripper_r", "label": "gripper_aperture",    "distance_type": "centroid", "axis": null, "plane": null}
  ],
  "sub_goal_constraints": [
    {
      "label": "object_lifted_sg",
      "description": "Object centroid must be at least 5 cm above the table (entity 3) centroid.",
      "expression": "lambda ee, keypoints, ee_R: 0.05 - (keypoints[1][2] - keypoints[3][2])",
      "reasoning": "Confirms the object has been lifted off the table by comparing against the current table height."
    }
  ]
},
{
  "subtraj_idx": 2,
  "reasoning": "ST2 transports the grasped object to the target zone. The gripper must remain closed throughout transport to avoid dropping the object.",
  "pairs": [
    {"from_entity": 1, "to_entity": 2, "label": "object_to_target_xy", "distance_type": "centroid", "axis": null, "plane": [0, 0, 1]},
    {"from_entity": 1, "to_entity": 2, "label": "object_to_target",    "distance_type": "centroid", "axis": null, "plane": null}
  ],
  "sub_goal_constraints": [
    {
      "label": "object_at_target_sg",
      "description": "Object centroid must be within 5 cm of the target zone.",
      "expression": "lambda ee, keypoints, ee_R: np.linalg.norm(keypoints[1] - keypoints[2]) - 0.05",
      "reasoning": "Object has reached the target location."
    }
  ]
}

Return JSON with the following schema:
{
  "subtrajectories": [
    {
      "subtraj_idx": <int or null>,
      "reasoning": <string>,
      "pairs": [
        {
          "from_entity": <int or "ee" or "gripper_l" or "gripper_r">,
          "to_entity": <int or "ee" or "gripper_l" or "gripper_r">,
          "label": <string>,
          "distance_type": <"centroid"|"surface"|"orientation"|"gripper_aperture">,
          "axis": <[x,y,z] or null>,
          "plane": <[nx,ny,nz] or null>,
          "ee_vector": <[x,y,z] or null>
        },
        ...
      ],
      "sub_goal_constraints": [
        {
          "label": <string>,
          "description": <string>,
          "expression": <python lambda string>,
          "reasoning": <string>
        },
        ...
      ]
    },
    ...
  ]
}

Only include clearly relevant pairs and constraints for each specified sub-trajectory.
\end{lstlisting}

\paragraph{User message.}
\begin{lstlisting}[style=transcript]
Task: pick up the alphabet soup and place it in the basket

Available objects (id: name):
  1: alphabet_soup_1
  2: basket_1
  3: salad_dressing_1
  4: cream_cheese_1
  5: milk_1
  6: tomato_sauce_1
  7: butter_1

Number of sub-trajectories: 2
  ST0: timesteps 0–61
  ST1: timesteps 62–163
\end{lstlisting}

\noindent The three images below accompany the user message above.

\paragraph{Assistant response.}
\begin{lstlisting}[style=transcript]
{
  "subtrajectories": [
    {
      "subtraj_idx": 0,
      "reasoning": "ST0 is the approach-and-grasp phase for the alphabet soup can. The most informative signals are the distances from each gripper finger to the can, the lateral alignment of the gripper with the can, and the gripper orientation for a top-down grasp.",
      "pairs": [
        {
          "from_entity": "gripper_l",
          "to_entity": 1,
          "label": "finger_l_to_alphabet_soup",
          "distance_type": "surface",
          "axis": null,
          "plane": null,
          "ee_vector": null
        },
        {
          "from_entity": "gripper_r",
          "to_entity": 1,
          "label": "finger_r_to_alphabet_soup",
          "distance_type": "surface",
          "axis": null,
          "plane": null,
          "ee_vector": null
        },
        {
          "from_entity": "gripper_l",
          "to_entity": 1,
          "label": "finger_l_to_alphabet_soup_xy",
          "distance_type": "centroid",
          "axis": null,
          "plane": [
            0,
            0,
            1
          ],
          "ee_vector": null
        },
        {
          "from_entity": "gripper_l",
          "to_entity": 1,
          "label": "finger_l_to_alphabet_soup_z",
          "distance_type": "centroid",
          "axis": [
            0,
            0,
            1
          ],
          "plane": null,
          "ee_vector": null
        },
        {
          "from_entity": "ee",
          "to_entity": "ee",
          "label": "gripper_down_alignment",
          "distance_type": "orientation",
          "axis": [
            0,
            0,
            -1
          ],
          "plane": null,
          "ee_vector": null
        }
      ],
      "sub_goal_constraints": [
        {
          "label": "fingers_near_alphabet_soup_sg",
          "description": "Both finger tips must be within 8 cm of the alphabet soup centroid at the end of the grasp approach.",
          "expression": "lambda ee, keypoints, ee_R: np.maximum(np.linalg.norm(keypoints['gripper_l'] - keypoints[1]), np.linalg.norm(keypoints['gripper_r'] - keypoints[1])) - 0.08",
          "reasoning": "This ensures the gripper has reached grasping proximity to the can before transitioning to transport."
        },
        {
          "label": "gripper_pointing_down_sg",
          "description": "The gripper approach direction must be within about 30 degrees of straight down at the end of ST0.",
          "expression": "lambda ee, keypoints, ee_R: 0.85 - np.dot(ee_R[:, 2], np.array([0.0, 0.0, -1.0]))",
          "reasoning": "A downward-facing gripper is the appropriate orientation for securely grasping the upright can from above."
        }
      ]
    },
    {
      "subtraj_idx": 1,
      "reasoning": "ST1 transports the grasped alphabet soup to the basket and lowers it into the basket region. The key signals are the can-to-basket lateral alignment, overall proximity to the basket, vertical relation to the basket, and maintaining a closed grasp during transport.",
      "pairs": [
        {
          "from_entity": 1,
          "to_entity": 2,
          "label": "alphabet_soup_to_basket_xy",
          "distance_type": "centroid",
          "axis": null,
          "plane": [
            0,
            0,
            1
          ],
          "ee_vector": null
        },
        {
          "from_entity": 1,
          "to_entity": 2,
          "label": "alphabet_soup_to_basket",
          "distance_type": "centroid",
          "axis": null,
          "plane": null,
          "ee_vector": null
        },
        {
          "from_entity": 1,
          "to_entity": 2,
          "label": "alphabet_soup_to_basket_z",
          "distance_type": "centroid",
          "axis": [
            0,
            0,
            1
          ],
          "plane": null,
          "ee_vector": null
        },
        {
          "from_entity": "gripper_l",
          "to_entity": "gripper_r",
          "label": "gripper_aperture",
          "distance_type": "gripper_aperture",
          "axis": null,
          "plane": null,
          "ee_vector": null
        }
      ],
      "sub_goal_constraints": [
        {
          "label": "alphabet_soup_aligned_with_basket_sg",
          "description": "The alphabet soup centroid must be horizontally within 10 cm of the basket centroid at the end of placement.",
          "expression": "lambda ee, keypoints, ee_R: np.linalg.norm((keypoints[1] - keypoints[2])[:2]) - 0.10",
          "reasoning": "This ensures the can is positioned over the basket opening rather than beside it."
        },
        {
          "label": "alphabet_soup_lowered_to_basket_sg",
          "description": "The alphabet soup centroid must be no more than 12 cm above the basket centroid at the end of placement.",
          "expression": "lambda ee, keypoints, ee_R: (keypoints[1][2] - keypoints[2][2]) - 0.12",
          "reasoning": "This confirms the can has been lowered into the basket region instead of remaining high above it."
        }
      ]
    }
  ]
}
\end{lstlisting}

\end{document}